\documentclass{ecai} 

\usepackage{latexsym}
\usepackage{color}
\usepackage{times}
\usepackage{soul}
\usepackage{url}
\usepackage[hidelinks]{hyperref}
\usepackage[utf8]{inputenc}
\usepackage[small]{caption}
\usepackage{graphicx}
\usepackage{amsmath}
\usepackage{amsthm}
\usepackage{booktabs}
\usepackage{algorithm}
\usepackage{algorithmic}
\usepackage[switch]{lineno}
\usepackage{enumerate}
\usepackage{multirow}
\usepackage{makecell}
\usepackage{threeparttable}
\usepackage{colortbl}
\usepackage{amssymb}
\usepackage{pifont}
\usepackage{xcolor}
\newtheorem{assumption}{Assumption}
\usepackage{latexsym}
\usepackage{marvosym}

\newtheorem{theorem}{Theorem}
\newtheorem{lemma}[theorem]{Lemma}

\newtheorem{proposition}[theorem]{Proposition}

\newcommand{\BibTeX}{B\kern-.05em{\sc i\kern-.025em b}\kern-.08em\TeX}

\begin{document}


\begin{frontmatter}


\paperid{123} 


\title{H$^2$Tune: Federated Foundation Model Fine-Tuning with Hybrid Heterogeneity}


\author[A]{\snm{$\text{Wei Guo}^{\dag}$}}
\author[B]{\snm{$\text{Siyuan Lu}^{\dag}$}}
\author[C]{\snm{Yiqi Tong}} 
\author[D]{\snm{Zhaojun Hu}} 
\author[A,E]{\snm{$\text{Fuzhen Zhuang}^{*}$}} 
\author[F]{\snm{$\text{Xiao Zhang}$}\thanks{Fuzhen Zhuang and Xiao Zhang are corresponding authors. Email: zhuangfuzhen@buaa.edu.cn, xiaozhang@sdu.edu.cn.\\ $\dag$ Equal contribution.}}
\author[G]{\snm{Tao Fan}} 
\author[H]{\snm{Jin Dong}} 

\address[A]{School of Artificial Intelligence, Beihang University}
\address[B]{School of Computer Science and Technology, Heilongjiang University}
\address[C]{School of Computer Science and Engineering, Beihang University}
\address[D]{School of Statistics, Renmin University of China}
\address[E]{Zhongguancun Laboratory, China}
\address[F]{School of Computer Science and Technology, Shandong University}
\address[G]{WeBank Co., Ltd, Shenzhen, China}
\address[H]{Beijing Academy of Blockchain and Edge Computing}

\begin{abstract}
Different from existing federated fine-tuning (FFT) methods for foundation models, hybrid heterogeneous federated fine-tuning (HHFFT) is an under-explored scenario where clients exhibit double heterogeneity in model architectures and downstream tasks. This hybrid heterogeneity introduces two significant challenges: \textit{\textbf{1) heterogeneous matrix aggregation}}, where clients adopt different large-scale foundation models based on their task requirements and resource limitations, leading to dimensional mismatches during LoRA parameter aggregation; and \textit{\textbf{2) multi-task knowledge interference}}, where local shared parameters, trained with both task-shared and task-specific knowledge, cannot ensure only task-shared knowledge is transferred between clients. To address these challenges, we propose \textit{H$^2$Tune}, a federated foundation model fine-tuning with hybrid heterogeneity. Our framework \textit{H$^2$Tune} consists of three key components: \textit{(i) sparsified triple matrix decomposition} to align hidden dimensions across clients through constructing rank-consistent middle matrices, with adaptive sparsification based on client resources; \textit{(ii) relation-guided matrix layer alignment} to handle heterogeneous layer structures and representation capabilities; and \textit{(iii) alternating task-knowledge disentanglement} mechanism to decouple shared and specific knowledge of local model parameters through alternating optimization. Theoretical analysis proves a convergence rate of $O(1/\sqrt{T})$. Extensive experiments show our method achieves up to 15.4\% accuracy improvement compared to state-of-the-art baselines. Our code is available at \hyperlink{https://anonymous.4open.science/r/H2Tune-1407}{https://anonymous.4open.science/r/H2Tune-1407}.
\end{abstract}

\end{frontmatter}



\section{Introduction}
Foundation models (FMs) integrated with federated learning (FL) \cite{yang2019federated} have emerged as a promising paradigm for optimizing client-specific applications \cite{kuang2024federatedscope,liu2023differentially}. This integration enables the customization of FMs to address downstream tasks through federated fine-tuning (FFT). Current research presents two primary methodologies for FFT: full-parameter fine-tuning (FPFT) \cite{han2024parameter} and parameter-efficient fine-tuning (PEFT) \cite{ding2023parameter}. The conventional FPFT approach updates all model parameters, which incurs prohibitive costs for FMs with billions of parameters on clients. In contrast, PEFT updates only a small subset of parameters while keeping others frozen, significantly reducing costs while maintaining comparable performance.
\begin{table}[t]
\centering
\caption{FFT framework comparisons with our proposed H$^2$Tune.}
\setlength{\tabcolsep}{0.7mm}{
\resizebox{0.47\textwidth}{!}{
\begin{tabular}{ccccccc}
\toprule
\multirow{3}{*}{Frameworks} & \multicolumn{6}{c}{Heterogeneity} \\ \cline{2-7}
                            & \multicolumn{4}{c}{Model}                & \multirow{2}{*}{Task} & \multirow{2}{*}{Resource} \\ \cline{2-5}
                            & Layer       & Dimension       & Architecture & LoRA        &                                     &                                         \\ \hline
FFTs                        & \ding{56} & \ding{56} & \ding{56}  & \ding{56} & \ding{56}                         & \ding{56}                             \\
HetLoRA \cite{cho2024heterogeneous}                     & \ding{56} & \ding{56} & \ding{56}  & \ding{52} & \ding{56}                         & \ding{52}                             \\
FlexLoRA \cite{bai2024federated}                    & \ding{56} & \ding{56} & \ding{56}  & \ding{52} & \ding{52}                         & \ding{56}                             \\
HeteroTune \cite{jia2024towards}                 & \ding{56} & \ding{52} & \ding{56}  & \ding{56} & \ding{56}                         & \ding{56}                             \\
pFedLoRA \cite{yi2023fedlora}                   & \ding{56} & \ding{52} & \ding{56}  & \ding{56} & \ding{56}                         & \ding{56}                             \\
\textbf{H$^2$Tune  }        & \color{red}\ding{52} & \ding{52} & \color{red}\ding{52}  & \ding{52} & \ding{52}                         & \ding{52}                             \\ 
\bottomrule
\end{tabular}
}
}
\label{tab:method comparisons}
\end{table}

Despite these advances, effectively handling task heterogeneity across clients remains a core challenge in FFT. Previous works have explored task heterogeneity through diverse approaches. FedBone \cite{chen2024fedbone} addresses this challenge by separating general and task-specific models via server-client split learning, while FedMSplit \cite{chen2022fedmsplit} leverages dynamic graphs to capture inter-task correlations. Other approaches like FedLPS \cite{jia2024fedlps} addresses task heterogeneity by sharing a general encoder across tasks while using adaptive pruning and heterogeneous aggregation for different client resources. These approaches work effectively for small models by identifying and sharing common layers while allowing personalized remaining layers. 
However, these approaches aren't directly applicable to large-scale FMs where clients often aggregate PEFT parameters. This creates two problems: clients with different tasks may fine-tune different model components, making parameter alignment difficult; and even when the same components are tuned, the parameters contain entangled task-shared and task-private knowledge. Current federated FM frameworks cannot effectively align diverse fine-tuning structures or decouple this knowledge, limiting effective transfer across heterogeneous tasks.

Additionally, in practical FFT scenarios, clients may not only select varying model components for PEFT. Still, they might also select different FM scales or model families with different architectures based on their task or resource constraints. We term this scenario model heterogeneous federated fine-tuning. Traditional model heterogeneous approaches in FL, like knowledge distillation-based \cite{zhang2022fine,chang2019cronus,cheng2021fedgems,zhang2022fedzkt,ahn2019wireless,ahn2020cooperative} or mutual learning-based methods \cite{shen2020federated,wu2022communication}, become infeasible due to the enormous computational and communication costs associated with FMs' parameter counts. Although model split-based methods \cite{jang2022fedclassavg,chen2021fedmatch,alam2022fedrolex,diao2020heterofl,horvath2021fjord} achieve partial parameter sharing by splitting client models into shared and private parts, however, it strictly assume shared homogeneous models. Recent works in federated foundation models have begun addressing this heterogeneity. As shown in Table \ref{tab:method comparisons}, HetLoRA \cite{cho2024heterogeneous} first attempted to address the challenge of heterogeneous LoRA aggregation with different ranks in FFT due to resource constraints through simple padding. FlexLoRA \cite{bai2024federated} advances this approach by distributing global LoRA parameters to client-specific ranks via SVD decomposition based on local requirements. These methods still require clients to apply LoRA to models from the same family with identical layer structures and dimensions. Although current works, HeteroTune \cite{jia2024towards} and pFedLoRA \cite{yi2023fedlora}, try to support heterogeneous hidden state dimensions, they also require uniform layer counts and model architecture.

\begin{figure}[t]
  \centering
  \includegraphics[scale=0.7]{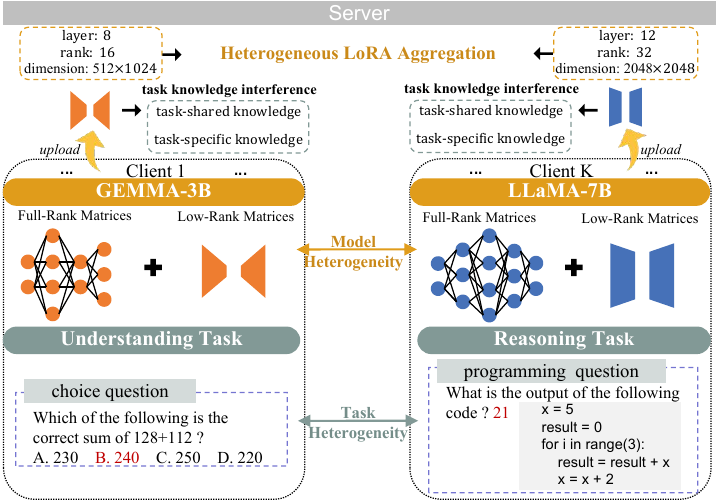}
  \caption{Challenges encountered in our proposed \textit{H$^2$Tune} framework. \textit{H$^2$Tune} focuses on hybrid heterogeneous FFT scenarios with large-scale foundation model heterogeneity and task heterogeneity. We address model heterogeneity across multiple levels, including layers, hidden dimensions, and model architectures, while solving the problem of inseparable task-shared and task-private knowledge in current federated fine-tuning frameworks.} 
  \label{fig:introduction}
  \vspace{14pt}  
\end{figure}

In this work, we identify an under-explored real-world scenario: hybrid heterogeneous federated fine-tuning (HHFFT), where clients simultaneously face task heterogeneity, resource heterogeneity, and model heterogeneity. For instance, in the healthcare field, large hospitals deploy LLaMA-7B \cite{touvron2023llama} for radiology report generation and disease progression prediction, while smaller clinics use Qwen-1.8B \cite{bai2023qwen} for medication recommendation and appointment scheduling. Similarly, in the financial sector, large investment banks leverage LLaMA-7B for trading strategy generation, while local credit unions adopt LLaMA-3B for credit scoring. In summary, we identify the following key challenges of HHFFT, as shown in Figure \ref{fig:introduction}:
\begin{enumerate}[0]
    \item[$\bullet$] \textit{\textbf{Heterogeneous matrix aggregation.}} It refers that clients may fine-tune different model components or use different model scales or architectures based on their needs, creating fine-tuned weight matrix alignment challenges during aggregation due to mismatched layers, incompatible dimensions, and varied architectures.
    \item[$\bullet$] \textit{\textbf{Multi-task knowledge interference.}} It refers to existing FFT methods that combine both shared and task-specific knowledge fail to prevent task-specific information from leaking into shared parameters. Given the task differences across clients, any knowledge specific to one task may interfere with other task performances.
\end{enumerate}

To address these challenges, we propose \textit{H$^2$Tune}, a federated foundation model fine-tuning framework with hybrid heterogeneity. \textit{H$^2$Tune} contains three key components: 1) \textit{sparsified triple matrix decomposition} that aligns hidden dimensions (width) across clients by constructing consistent shared middle matrices, with adaptive sparsification based on each client's resources; 2) \textit{relation-guided matrix layer alignment} that enables heterogeneous layer (depth) aggregation by learning cross-layer matrix relationships via relation networks; and 3) \textit{alternating task-knowledge disentanglement} that leverage alternating optimization to decouple local models into shared matrices and private matrices, and achieve federated knowledge transfer by shared matrices with task-shared knowledge. Our main contributions are summarized as follows:
\begin{enumerate}[\hspace{0em}(a)]
    \item We identify an under-explored scenario in FFT named hybrid heterogeneous federated fine-tuning, where task, model, and resource heterogeneity coexist. This hybrid heterogeneity introduces unique challenges beyond current FFT methods. \textit{H$^2$Tune} is developed to handle heterogeneity in task objectives, model structures, and resource constraints within FFT.
    \item \textit{H$^2$Tune} is carefully designed with two core components: 1) a sparse triple matrix decomposition with relation-guided alignment for heterogeneous matrix alignment across clients, and 2) an alternating disentanglement optimization mechanism that separates general and task-specific knowledge.
    \item Our approach accommodates both homogeneous and heterogeneous model settings and supports single or multi-task scenarios. And we theoretically establish the convergence guarantee of \textit{H$^2$Tune}, proving that it achieves an $O(1/\sqrt{T})$ convergence rate.
    \item Extensive experiments on two well-known benchmarks validate \textit{H$^2$Tune} outperforms state-of-the-art baselines by up to 15.4\% accuracy in both homogeneous and heterogeneous scenarios.
\end{enumerate}

\begin{figure*}[t]
  \centering
  \includegraphics[scale=0.78]{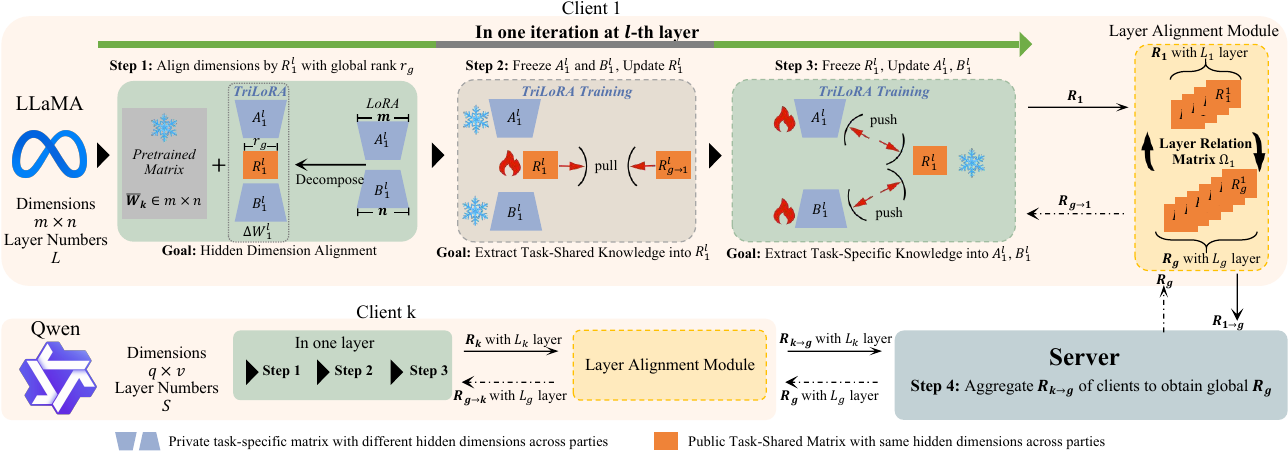}
  \caption{Overview of one client in \textit{H$^2$Tune} in one communication round. Blue regions indicate task-specific knowledge, orange for task-shared knowledge, and white for sparsified parameters. Clients align matrices through triple factorization and relation matrices for global intermediate matrices, with sparsity adaptation to local resources. Alternating optimization decouples shared and specific knowledge, enabling cross-client fine-tuning through exchanging intermediate matrices containing task-shared knowledge.} 
  \label{fig: framework}
  \vspace{14pt}  
\end{figure*}

\section{Related Work}
Foundation models (FMs) have been widely adopted in federated learning, with fine-tuning approaches primarily divided into full-parameter fine-tuning (FPFT) \cite{han2024parameter} and parameter-efficient fine-tuning (PEFT) \cite{ding2023parameter}. While FPFT offers superior performance, its computational and communication costs are substantial \cite{liu2024hift}. In contrast, PEFT methods like LoRA \cite{hu2021lora} provide efficient alternatives by updating only partial parameters, showing promising results in federated scenarios \cite{jiang2024low,zhang2024towards}. However, multi-task federated fine-tuning for FMs remains challenging. Recent work \cite{bai2024federated,chen2022fedmsplit,chen2024fedbone} have made initial progress, but limitations persist. They either lack PEFT support for computational efficiency, or fail to effectively distill task-shared knowledge for cross-client transfer. More critically, they require identical foundation models across clients.

Model heterogeneity in FL has attracted substantial research attention. These works can be categorized into two main branches. Specifically, partially model heterogeneous methods allow clients to maintain different subnets of a global architecture that can be aggregated at the server, as demonstrated in FedRolex \cite{alam2022fedrolex}, HeteroFL \cite{diao2020heterofl}, FjORD \cite{horvath2021fjord}, and HFL \cite{lu2021heterogeneous}. On the other hand, the completely model-heterogeneous branch involves clients with entirely different model structures that cannot be directly aggregated, and is further divided into the following three categories: knowledge distillation-based methods that either use public datasets \cite{chang2019cronus,cheng2021fedgems} or employ alternative approaches like FedZKT \cite{zhang2022fedzkt} which generates synthetic datasets through generator training, and HFD \cite{ahn2019wireless,ahn2020cooperative} which requires uploading local logits for server aggregation, all of which exacerbate substantial computational overhead for large-scale FMs in FL; model split-based methods that share either feature extractors \cite{chen2021fedmatch} or classifiers \cite{jang2022fedclassavg} while experiencing performance bottlenecks from partial parameter sharing; and mutual learning-based methods that assign small homogeneous and large heterogeneous models to each client \cite{shen2020federated,wu2022communication}, yet fail to optimize the relationship between model structure and parameter capacity.

However, in federated FM fine-tuning scenarios, the massive parameter space of FMs further amplifies the limitations of these existing approaches. Although PEFT techniques reduce update parameters for large FMs, they introduce a critical new challenge: clients may deploy PEFT methods like LoRA or Adaptor across different architectural layers or foundation model families, creating a novel form of model heterogeneity: model heterogeneous federated fine-tuning. Recent research has addressed LoRA heterogeneity within identical model frameworks: HetLoRA \cite{cho2024heterogeneous} mitigates rank heterogeneity through parameter alignment via simple padding, while FlexLoRA \cite{bai2024federated} employs SVD decomposition to distribute global parameters to client-specific ranks for personalized adaptation. Despite HeteroTune \cite{jia2024towards} and pFedLoRA \cite{yi2023fedlora} attempting to resolve hidden dimension heterogeneity, these approaches still remain constrained to uniform model families with same architectural configurations. Moreover, considering that clients in real-world scenarios often have diverse tasks and resource constraints, these approaches similarly overlook the challenges posed by such practical heterogeneity.

\section{Preliminary}
\textbf{Federated fine-tuning with hybrid heterogeneity.} A HHFFT framework coordinates $K$ distributed clients with a central server, where each client $k$ maintains a private dataset $\mathcal{D}_k = \{(x_i,y_i), i= 1,\cdots,N_k\}$ and a foundation model $\mathbf{W}_k\subseteq \mathbb{R}^d$. By sharing parameters $\Delta\mathbf{W}_k \subseteq \mathbb{R}^q(q << d)$, HHFFT enables collaborative learning across heterogeneous environments, allowing clients with different foundation models, resource constraints, and tasks to enhance their individual model performance by $\mathbf{W}_k = \bar{\mathbf{W}}_k+\Delta\mathbf{W}_k$, where $\bar{\mathbf{W}}_k$ denotes the frozen parameters of client $k$. The fine-tuning parameters $\Delta\mathbf{W}_k$ can be represented as $\Delta\mathbf{W}_k= \{W^l_k\}_{l=1}^{L_k}$, where $L_k$ denotes the number of layers and $W^l_k \subseteq \mathbb{R}^{d_{\text{in}}^k \times d_{\text{out}}^k}$ represents the $l$-th layer parameters with input and output dimensions $d_{\text{in}}^k$ and $d_{\text{out}}^k$. Each client $k$ optimizes its task-specific function $\mathcal{L}_k$ over local dataset $\mathcal{D}_k$.


Formally, we define $\mathcal{M}$ as the foundation model, $\mathcal{T}$ as the downstream task, and $\mathcal{R}$ as computational resources. For any two clients $k,j \in \{1,...,K\} (k \neq j)$, the heterogeneous settings in HHFFT manifest: model heterogeneity ($\mathcal{M}_k \neq \mathcal{M}_j$), task heterogeneity ($\mathcal{T}_k \neq \mathcal{T}_j$), and resource heterogeneity ($\mathcal{R}_k \neq \mathcal{R}_j$). Notably, model heterogeneity in FFT primarily manifests as heterogeneity in the $\{\Delta\mathbf{W}_k\}_{k=1}^{K}$ being aggregated, which encompasses four key aspects:
\begin{itemize}
 \item[$\blacktriangleright$] \textit{Rank heterogeneity:} $r_k \neq r_j$,
 \item[$\blacktriangleright$] \textit{Dimensional heterogeneity:} $d_{\text{in}}^k \neq d_{\text{in}}^j$ or $d_{\text{out}}^k \neq d_{\text{out}}^j$,
 \item[$\blacktriangleright$] \textit{Layer heterogeneity:} $L_k \neq L_j$,
 \item[$\blacktriangleright$] \textit{Architecture heterogeneity:} $F_k \neq F_j$,
\end{itemize}
where $r_k$ refers to the intermediate dimensions of low-rank fine-tuning matrix, with each layer-wise update $\Delta W_k^l = A_k^l \times B_k^l$, where $A_k^l \in \mathbb{R}^{d_{\text{in}} \times r_k}$ and $B_k^l \in \mathbb{R}^{r_k \times d_{\text{out}}}$ and $(r_k<<\text{min}(d_{\text{in}}, d_{\text{out}}))$.

\noindent\textbf{Problem Formulation.} The optimization objective of \textit{H$^2$Tune} is:
\begin{equation}
\begin{aligned}
    &\mathop{\rm min}_{\Delta\mathbf{W}_1,\,\ldots,\,\Delta\mathbf{W}_K} \sum_{k=1}^{K} \mathcal{L}_{k}(\Delta \mathbf{W}_k;\, \mathcal{M}_k,\, \mathcal{T}_k,\, \mathcal{R}_k), \\
    &\text{s.t.}~ \mathcal{M}_k \neq \mathcal{M}_j,~ \mathcal{T}_k \neq \mathcal{T}_j,~ \mathcal{R}_k \neq \mathcal{R}_j,~ \forall k \neq j.
\end{aligned}
\end{equation}

\section{Methodology}
\subsection{Overview}
\begin{figure*}[t]
  \centering
  \includegraphics[scale=0.88]{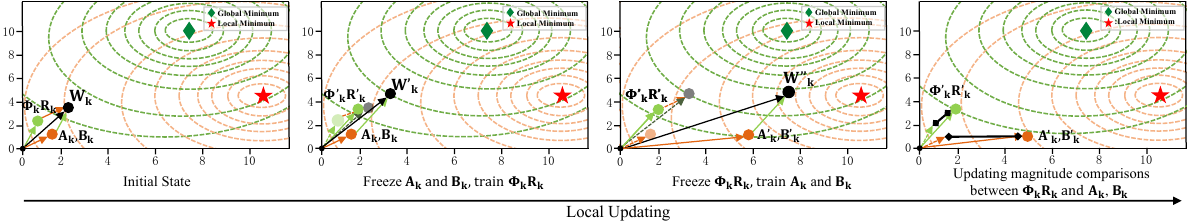}
  \caption{A toy example to illustrate the alternating task-knowledge disentanglement. Private matrices $\mathbf{A}_k$ and $\mathbf{B}_k$ optimize toward local optima with task-specific knowledge, while shared matrix $\mathbf{R}_k$ optimizes toward the global optimum with shared knowledge. Due to their larger parameter number, $\mathbf{A}_k$ and $\mathbf{B}_k$ dominate the optimization direction toward local optima, while independent updates between private and shared matrices ensure optimization order does not affect the model optimization direction.} 
  \label{fig: task decompsition}
  \vspace{4pt}  
\end{figure*}

We first provide an overview of $\textit{H$^2$Tune}$. As illustrated in Figure \ref{fig: framework}, each layer parameter matrix $W_k^l$ of client $k$ is decomposed into a frozen pre-trained matrix $\bar{W}_k^l$ plus a low-rank fine-tuning matrix $\Delta W_k^l$. Different from LoRA in PEFT which decomposes $\Delta W_k^l$ into two components: $A_k^l \subseteq \mathbb{R}^{a\times r_k}$ and $B_k^l\subseteq \mathbb{R}^{r_k\times b}$, we use three components: $A_k^l \in \mathbb{R}^{a\times r_k}$, $R_k^l \in \mathbb{R}^{r_k\times r_k}$, and $B_k^l \in \mathbb{R}^{r_k\times b}$, where $R^l_k$ is a dense real matrix rather than a diagonal intermediate matrix as in SVD. The training process in each communication round involves: 1) each client $k$ freezes its full-rank matrices $\mathbf{\bar{W}}_k$ and task-shared matrix $\mathbf{R}_k$, while updating the task-specific matrices $\mathbf{A_k}$ and $\mathbf{B}_k$. 2) each client $k$ aligns the layers between received global task-shared matrices $\mathbf{R}_g$ and local task-shared matrices using the layer relation matrix $\Omega_k$. Client then updates $\mathbf{R}_k$ and $\Omega_k$ while keeping task-specific matrices $\mathbf{A}_k$ and $\mathbf{B}_k$ frozen. After local updating, all clients only upload their task-shared matrices $\mathbf{R}_k$ to server, maintaining privacy of their task-specific matrices. 3) server receives clients' task-shared matrices and aggregates them to generate global task-shared matrices $\mathbf{R}_g$. Once the aggregation is finished, server sends $\mathbf{R}_g$ back to all clients. 4) Each client receives the global task-shared matrices and achieves local optimization. For notational simplicity, bold symbols denote collections across all $L_k$ layers: $\mathbf{W}_k = \{W_k^l\}_{l=1}^{L_k}$, $\mathbf{\bar{W}_k}= \{\bar{W}_k^l\}_{l=1}^{L_k}$, $\Delta\mathbf{W}_k =\{\Delta W_k^l\}_{l=1}^{L_k}$, $\mathbf{A}_k = \{A_k^l\}_{l=1}^{L_k}$, $\mathbf{B}_k = \{B_k^l\}_{l=1}^{L_k}$, $\mathbf{R}_k = \{R_k^l\}_{l=1}^{L_k}$, $\mathbf{\Phi}_k = \{\Phi_k^l\}_{l=1}^{L_k}$ and $\mathbf{R}_g = \{R_g^l\}_{l=1}^{L_g}$, where $L_g$ represents the layer number of global task-shared matrix $\mathbf{R}_g$.

\subsection{Sparsified Triple Matrix Decomposition}
To address hidden dimension heterogeneity, we introduce TriLoRA, which decomposes each layer's parameters into private task-specific matrices $A_k^l$, $B_k^l$ and a public task-shared matrix $R_k^l$. Though clients have varying computational resources that would normally lead to different local ranks $r_k$, we maintain a uniform global rank $r_g$ across all task-shared matrices for effective knowledge sharing. We accommodate resource differences through client-specific sparsity rates $\beta_k$ in sparse matrix $\phi_k^l$, adjusting practical parameter update. Specifically, for each client $k$, we decompose local FM $\mathbf{W}_k$ as:
\begin{equation}
    \mathbf{W}_k = \bar{\mathbf{W}}_k + \Delta \mathbf{W}_k.
\end{equation}
We decompose each layer's weight matrix $\Delta W_k^l$ of local model $\Delta \mathbf{W}_k$ into two private task-specific matrices $A_k^l$, $B_k^l$ and one public task-shared matrix $R_k^l$, combining them via residual connection:
\begin{equation}
\begin{aligned}
    \Delta W_k^l = &\big(A_k^l + A_k^l \cdot (\Phi_k^l\cdot R_k^l)\big) B_k^l,\\
    \text{s.t.} \ A_k^l \in \mathbb{R}^{a_k^l\times r_g}, &\ R_k^l,\Phi_k^l \in \mathbb{R}^{r_g \times r_g}, \ B_k^l \in \mathbb{R}^{r_g\times b_k^l},
\end{aligned}
\end{equation}
where $a_k^l$ and $b_k^l$ represent input and output dimensions at layer $l$. The global rank $r_g$ satisfies $0 < r \leq \min_{k\in[1,K]}(a_k^l,b_k^l)$, ensuring compatibility across clients with heterogeneous model dimensions. The sparsification matrix $\Phi_k^l$ has client-specific sparsity ratio $\beta_k \in [0,1]$, controlling the proportion of non-zero elements and allowing clients to adjust updatable parameters based on available resources. 


\subsection{Relation-Guided Matrix Layer Alignment}
Due to model scale or architectural heterogeneity across clients, clients have varying numbers of layers $L_k$ in their shared matrices $\mathbf{R}_k \in \mathbb{R}^{r_g\times r_g \times L_k}$, preventing direct server aggregation. To solve this layer mismatch, we introduce a trainable layer relation alignment matrix $\Omega_k \in \mathbb{R}^{L_k \times L_g}$ for each client $k$, where $L_g = \max(L_{k},\ k\in[1,K])$ represents the global layer count. The matrix $\Omega_k$ transforms local task-shared matrices $\mathbf{R}_k$ into uniform-sized representations for server aggregation:
\begin{equation}
\begin{aligned}
    &\mathbf{R}_{k \rightarrow g} := \mathbf{R}_{k}\Omega_{k}, \ \Omega_{k} \in \mathbb{R}^{L_{k}\times L_g}. \\
\end{aligned}
\end{equation}
The server averages these aligned matrices $\{\mathbf{R}_{k \rightarrow g}\}_{k=1}^K$ to produce the global task-shared matrix $\mathbf{R}_g \in \mathbb{R}^{r_g\times r_g \times L_g}$. Upon receiving $\mathbf{R}_g$, each client use $\Omega_k^\top$ to transform it into a locally compatible form:
\begin{equation}
    \mathbf{R}_{g \rightarrow k} := \mathbf{R}_{g}\Omega_{k}^\top, \ \Omega_{k}^\top \in \mathbb{R}^{L_g \times L_{k}}.
\end{equation}
These transformed matrices $\mathbf{R}_{g \rightarrow k}$ and $\Omega_{k}$ are further personalized to local distributions through loss $\mathcal{L}_{\rm share}^k$ in Equation \ref{equ:share loss}.

\subsection{Alternating Task-Knowledge Disentanglement}
\label{sec:alter}
In existing task heterogeneous FFT methods \cite{bai2024federated,chen2022fedmsplit,chen2024fedbone}, parameters are jointly trained with both task-specific and task-shared knowledge. This joint training causes the shared parameters to be contaminated by task-specific knowledge, hindering cross-client knowledge transfer. To address this interference, we propose an alternating optimization approach that separates task-shared knowledge into matrix $\mathbf{R}_k$ and task-specific knowledge into matrices $\mathbf{A}_k$ and $\mathbf{B}_k$. In each communication round $t\in [1,T]$, local iterations $e \in [1,E]$ consist of two key optimization steps.

First, client $k$ receives the global task-shared matrix $\mathbf{R}_{g}$ and transforms it to local form $\mathbf{R}_{g \rightarrow k}$ by $\Omega_{k}^\top$. Then, while freezing $\mathbf{R}_{g \rightarrow k}$, $\mathbf{A}_k$, and $\mathbf{B}_k$, the client optimizes $\mathbf{R}_{k}$, sparse matrix $\mathbf{\Phi}_k$, and layer relation matrix $\Omega_k$ by minimizing:
\begin{equation}
    \mathcal{L}_{\rm share}^k = \text{CE}(y^{\prime}_i , y_i) + \text{KL}(\mathbf{R}_k, \mathbf{R}_{g \rightarrow k}),
    \label{equ:share loss}
\end{equation}
where $y^{\prime}_i$ is the predicted label and CE is the cross-entropy loss.

In the second step, to further disentangle shared and private parameters, each client freezes $\mathbf{R}_k$, $\mathbf{\Phi}_k$, and $\Omega_k$, and optimizes task-specific matrices $\mathbf{A}_k$ and $\mathbf{B}_k$ by:
\begin{equation}
\mathcal{L}_{\text{specific}}^k = \text{CE}( y^{\prime\prime}_i , y_i) - \text{KL}(y^{\prime\prime}_i, y^{\prime}_i)\\
+\frac{\mathcal{V}}{2}(\Vert \mathbf{A}_k\Vert_2+\Vert \mathbf{B}_k\Vert_2),
\end{equation}
where $y^{\prime\prime}_i$ is the prediction based on optimized $\mathbf{R}_k$, and $\mathcal{V}$ is the regularization coefficient. After local updating, each client uploads only $\mathbf{R}_k$ to server while keeping $\mathbf{A}_k$ and $\mathbf{B}_k$ private.

\textbf{Toy example.} Figure \ref{fig: task decompsition} illustrates our local two-stage update process. The green $\blacklozenge$ represents the optimal point for the task-shared model $\mathbf{R_k}$ (trained on all clients' data), while the red $\bigstar$ represents the optimal point for task-specific matrices $\mathbf{A}_k$ and $\mathbf{B}_k$ (trained on individual client data). $\mathcal{L}_{\text{shared}}$ guides $\mathbf{R}_k$ toward global knowledge, while $\mathcal{L}_{\text{specific}}$ directs $\mathbf{A}_k$ and $\mathbf{B}_k$ toward client-specific knowledge. Since $\mathbf{R}_k$ and $\mathbf{A}_k, \mathbf{B}_k$ are updated independently, the optimization paths remain unbiased. However, as shown in the rightmost figure, $\mathbf{R}_k$'s lower dimensionality results in smaller update steps compared to $\mathbf{A}_k$ and $\mathbf{B}_k$. This approach ensures that local parameters primarily serve the local task while still benefiting from global knowledge.



\section{Convergence Analysis}\label{sec:conver}
In this section, we present theoretical guarantees for the convergence of our proposed \textit{H$^2$Tune}. In contrast to traditional alternating optimization, our approach involves different optimization targets and models in its two-step client optimization, forming a bidirectional optimization process. Let $\mathbf{R}:=\{\mathbf{R}_1,...,\mathbf{R}_K\}$ and $\mathbf{H} := \{(\mathbf{A}_1,\mathbf{B}_1),...,(\mathbf{A}_K,\mathbf{B}_K)\}$ represents task-shared parameters and task-specific parameters across all clients, respectively. We denote the complete set of parameters in the federated learning system as $\Theta := \{\mathbf{R}, \mathbf{H}\}$. Specifically, denote $L(\mathbf{R}, \mathbf{H}) = \sum_{k=1}^K \mathcal{L}_{\text{share}}^k(\mathbf{R}_k, \mathbf{H}_k)$, $h(\mathbf{R}) = \lambda \sum_{k=1}^K  KL(\mathbf{R}_k, \mathbf{R}_g)$, ${G}_0(\mathbf{R},$ $\mathbf{H}) = \sum_{k=1}^K \mathcal{L}_{\text{specific}}^k(\mathbf{R}_k, \mathbf{H}_k)$, and ${G}(\mathbf{R}, \mathbf{H}) = \sum_{k=1}^K \mathcal{L}_{\text{specific}}^k( \mathbf{R}_k, \mathbf{H}_k) + \frac {\mathcal{V}}{2} \sum_{k=1}^K \|\mathbf{H}_k\|^2$. Then, our optimization problem could be written as follows:
\begin{equation}
\begin{aligned}
    &\min_{\mathbf{R}} L(\mathbf{R}, \mathbf{H}^*(\mathbf{R}))+ h(\mathbf{R}),\\
    \text{s.t.} &\quad \mathbf{H}^*(\mathbf{R}) = \arg\min_{\mathbf{H}} {G}(\mathbf{R}, \mathbf{H}),
\end{aligned}
\end{equation}
where $F(\mathbf{R}) := L(\mathbf{R}, \mathbf{H}^*(\mathbf{R}))$. Let $\Theta := \{\mathbf{R}, \mathbf{H}\}$ denote all parameters. Then, we make some standard assumptions (\cite{ghadimi2018approximation,ji2020convergence,rajput2020closing}) on $L$ and ${G}$ of our bi-level optimization problem.

\begin{assumption}[Lipschitz Condition]
    The loss function $\mathcal{L}(\Theta)$ is $L_1$-Lipschitz.
    \label{ass 1}
\end{assumption}

\begin{assumption}[Smoothness]
    The loss function $\mathcal{L}(\Theta)$, $G(\Theta)$ and $G_0(\Theta)$ are $L_2$-smooth, $L_2$-smooth and $L$-smooth, respectively.
    \label{ass 2}
\end{assumption}

\begin{assumption}[Lipschitz Condition for Second Derivatives]
    The second derivatives $\nabla_\mathbf{R}\nabla_\mathbf{H}G(\Theta)$ and $\nabla_\mathbf{H}^2G(\Theta)$ are $L_3$-Lipschitz and $L_4$-Lipschitz, respectively.
    \label{ass 3}
\end{assumption}

\begin{assumption}[Bounded Domain]
    The parameter $\mathbf{H}$ is in a bounded domain with a diameter $\Delta$, i.e., for any $\mathbf{H}$ and $\mathbf{H}'$, we have:
    \begin{equation}
        \Vert \mathbf{H}-\mathbf{H'} \Vert \leq \Delta. 
    \end{equation}
    \label{ass 4}
\end{assumption}

\begin{assumption}
    Function $\mathcal{F} = F(\mathbf{R}) + h(\mathbf{R})$ is bounded below:
    \begin{equation}
        \mathcal{F}^{\ast} = \inf_{\mathbf{R}}\mathcal{F}(\mathbf{R})>-\infty.
    \end{equation}
    \label{ass 5}
\end{assumption}
\begin{theorem}\label{thm1}
Under Assumption \ref{ass 1} - \ref{ass 5}, define $\alpha = -L + \mathcal{V}$, choose step size $\eta$ to be $\frac{2}{L_2 + \alpha}$, $\eta^{\prime}$ to be $\frac{1}{6L_0}$, and suppose $\alpha < L_2$, $h(\mathbf{R}^{\tau-1}_t) \leq h(\mathbf{R}^0_t),\forall t$, we have:
\begin{equation}
    \frac{1}{\tau T}\sum^{T-1}_{t=0}\sum_{j=1}^{\tau-1}\Vert \mathcal{G}_t^{j}\Vert^2 \leq \mathcal{O}(\frac{1}{T}),
\end{equation}
where $\|\mathcal{G}_t^j\|$ is used to measure convergence properties of our algorithms. We define the generalized gradient at the $j$-th iteration of $t$-th round: $\mathcal{G}_t^j = \frac{1}{\eta'} \left(\mathbf{R}_t^j - \mathbf{R}_t^{j+1}\right)$. $\mathcal{V}$ refers to the regularization coefficient of $\mathcal{L}_{specific}$, and $\tau$ is the number of local epochs.
\end{theorem}

\begin{table}[t]
\centering
\caption{Experimental scenario settings for model heterogeneity.}
\renewcommand{\arraystretch}{0.875} 
\setlength{\tabcolsep}{4mm}{
\resizebox{0.45\textwidth}{!}{
\begin{tabular}{ccccc}
\toprule
Scenario  & Model    & Scale & Layer & Dimension \\ 
\midrule
\multirow{3}{*}{Scenario 1} 
          & Gemma    & 2B    & 18    & 2,048 \\ 
          & Llama3.2 & 3B    & 28    & 3,072 \\ 
          & SmolLM   & 1.7B  & 24    & 2,048 \\ 
\midrule
\multirow{3}{*}{Scenario 2} 
         & Llama3   & 8B    & 32    & 4,096 \\ 
         & Llama3.2 & 3B    & 28    & 3,072 \\ 
        & Llama3.2 & 1B    & 16    & 2,048 \\ 
\midrule
\multirow{3}{*}{Scenario 3} 
         & Gemma    & 7B    & 28    & 3,072 \\ 
         & Llama3   & 8B    & 32    & 4,096 \\ 
         & Yi       & 6B    & 32    & 4,096 \\ 
\bottomrule
\end{tabular}
}
}
\label{tab:models}
\end{table}

\begin{table*}[t]
\centering
\caption{The overall performance comparison results in different scenarios and task settings, where each client handles a distinct task and $@r$ denotes the rank of low-rank updates supported by the client's resource constraints.}
\renewcommand{\arraystretch}{0.875} 
\small
\resizebox{\textwidth}{!}{
\begin{tabular}{c|c|ccc|ccc|ccc|ccc}
\toprule
\multirow{4}{*}{Scenario} & \multirow{4}{*}{Model} & \multicolumn{6}{c|}{Homogeneous Task Scenario} & \multicolumn{6}{c}{Heterogeneous Task Scenario} \\ 
\cmidrule(lr){3-8} \cmidrule(lr){9-14}
 &  & \multicolumn{3}{c|}{MATHInstruct} & \multicolumn{3}{c|}{GLUE} & \multicolumn{3}{c|}{MATHInstruct} & \multicolumn{3}{c}{GLUE} \\ 
 \cmidrule(lr){3-5} \cmidrule(lr){6-8} \cmidrule(lr){9-11} \cmidrule(lr){12-14}
 &  & \small MCQ$_{@16}$ & \small MCQ$_{@64}$&\small MCQ$_{@128}$ & \small SSC$_{@16}$ & \small SSC$_{@64}$ & \small SSC$_{@128}$ &\small FIBQ$_{@16}$  &  \small MCQ$_{@64}$ &  \small PQ$_{@128}$ &  \small SSC$_{@16}$&  \small SPC$_{@64}$ & \small NLI$_{@128}$ \\ 
\midrule
\multirow{7}{*}{Scenario 1} 
& LOCAL & 26.0 & 30.6 & 17.2 & 93.8 & 93.6 & 92.2 & 28.8 & 32.0 & 12.2 & 90.0 & 83.8 & 54.2 \\
& FLLM & 26.4 & 32.2 & 18.0 & 94.2 & 93.6 & 92.4 & 29.4 & 35.6 & 14.4 & 90.6 & 86.2 & 77.6 \\
& HetLoRA & 27.0 & 33.6 & 18.2 & 94.4 & 93.8 & 92.4 & 29.4 & 29.0 & 12.2 & 95.6 & 87.0 & 86.8 \\  
& FLTLA & 24.2 & 21.0 & 16.8 & 93.6 & 92.8 & 91.8 & 23.2 & 27.8 & 7.8 & 94.0 & 84.4 & 64.0 \\
&\cellcolor{gray!30}H$^2$Tune &\cellcolor{gray!30}\textbf{29.2} &\cellcolor{gray!30}\textbf{34.4} &\cellcolor{gray!30}\textbf{20.6} &\cellcolor{gray!30}\textbf{94.4} &\cellcolor{gray!30}\textbf{94.2} &\cellcolor{gray!30}\textbf{92.6} &\cellcolor{gray!30}\textbf{32.4} &\cellcolor{gray!30}\textbf{35.0} &\cellcolor{gray!30}\textbf{17.2} &\cellcolor{gray!30}\textbf{95.8} &\cellcolor{gray!30}\textbf{87.0} &\cellcolor{gray!30}\textbf{86.0} \\
&\cellcolor{gray!20}\textit{avg. Imp} &\cellcolor{gray!20}\textit{\textbf{3.3$\uparrow$}} &\cellcolor{gray!20}\textit{\textbf{5.1$\uparrow$}} &\cellcolor{gray!20}\textit{\textbf{3.1$\uparrow$}} &\cellcolor{gray!20}\textit{\textbf{0.4$\uparrow$}} &\cellcolor{gray!20}\textit{\textbf{0.8$\uparrow$}} &\cellcolor{gray!20}\textit{\textbf{0.4$\uparrow$}} &\cellcolor{gray!20}\textit{\textbf{4.7$\uparrow$}} &\cellcolor{gray!20}\textit{\textbf{3.9$\uparrow$}} &\cellcolor{gray!20}\textit{\textbf{5.6$\uparrow$}} &\cellcolor{gray!20}\textit{\textbf{3.3$\uparrow$}} &\cellcolor{gray!20}\textit{\textbf{1.7$\uparrow$}} &\cellcolor{gray!20}\textit{\textbf{15.4$\uparrow$}} \\
& Upper Boundary & 34.8 & 37.8 & 28.2 & 95.4 & 95.4 & 95.2 & 34.8 & 36.2 & 18.2 & 97.6 & 89.2 & 87.2 \\ 
\midrule
\multirow{7}{*}{Scenario 2}    
& LOCAL & 36.4 & 30.6 & 23.6 & 81.8 & 93.6 & 72.2 & 53.2 & 32.0 & 3.2 & 52.2 & 83.8 & 81.4 \\ 
& FLLM & 37 & 31.2 & 24.4 & 83.6 & 93.6 & 79.0 & 56.4 & 32.2 & 7.2 & 52.4 & 88.2 & 85.0 \\ 
& HetLoRA & 40.8 & 32.0 & 24.4 & 84.8 & 93.6 & 80.2 & 57.0 & 33.0 & 8.8 & 53.0 & 88.4 & 85.2 \\
& FLTLA & 28.8 & 29.0 & 21.2 & 81.4 & 85.4 & 72.2 & 53.4 & 31.0 & 6.6 & 52.2 & 83.0 & 81.4 \\
&\cellcolor{gray!30}H$^2$Tune &\cellcolor{gray!30}\textbf{42.2} &\cellcolor{gray!30}\textbf{33.2} &\cellcolor{gray!30}\textbf{25.2} &\cellcolor{gray!30}\textbf{87.8} &\cellcolor{gray!30}\textbf{93.8} &\cellcolor{gray!30}\textbf{81.4} &\cellcolor{gray!30}\textbf{57.2} & \cellcolor{gray!30}\textbf{33.0} & \cellcolor{gray!30}\textbf{9.0} & \cellcolor{gray!30}\textbf{53.0} & \cellcolor{gray!30}\textbf{88.4} & \cellcolor{gray!30}\textbf{85.2} \\  
&\cellcolor{gray!20}\textit{avg. Imp} &\cellcolor{gray!20}\textit{\textbf{6.5$\uparrow$}} &\cellcolor{gray!20}\textit{\textbf{2.5$\uparrow$}} &\cellcolor{gray!20}\textit{\textbf{1.8$\uparrow$}} &\cellcolor{gray!20}\textit{\textbf{4.9$\uparrow$}} &\cellcolor{gray!20}\textit{\textbf{2.3$\uparrow$}} &\cellcolor{gray!20}\textit{\textbf{5.5$\uparrow$}} &\cellcolor{gray!20}\textit{\textbf{2.3$\uparrow$}} &\cellcolor{gray!20}\textit{\textbf{1.0$\uparrow$}} &\cellcolor{gray!20}\textit{\textbf{2.6$\uparrow$}} &\cellcolor{gray!20}\textit{\textbf{0.6$\uparrow$}} &\cellcolor{gray!20}\textit{\textbf{2.6$\uparrow$}} &\cellcolor{gray!20}\textit{\textbf{2.0$\uparrow$}} \\
& Upper Boundary & 43.6 & 37.8 & 27.4 & 90.8 & 95.4 & 87.2 & 57.6 & 39.0 & 12.0 & 61.0 & 90.4 & 86.0 \\ 
\midrule
\multirow{7}{*}{Scenario 3}       
& LOCAL & 30.8 & 41.4 & 32.4 & 87.8 & 83.6 & 94.4 & 57.0 & 45.0 & 5.6 & 94.2 & 83.6 & 89.8 \\ 
& FLLM & 32.0 & 41.4 & 33.0 & 87.8 & 84.8 & 94.6 & 57.6 & 45.4 & 10.4 & 95.2 & 86.2 & 89.8 \\   
& HetLoRA & 35.8 & 43.2 & 36.0 & 92.4 & 87.0 & 95.4 & 57.6 & 45.8 & 9.6 & 96.0 & 87.6 & 90.4 \\
& FLTLA & 30.4 & 40.2 & 31.0 & 82.2 & 83.0 & 93.2 & 55.8 & 44.0 & 6.0 & 94.8 & 84.6 & 88.8 \\
& \cellcolor{gray!30}H$^2$Tune & \cellcolor{gray!30}\textbf{36.0} & \cellcolor{gray!30}\textbf{43.6} & \cellcolor{gray!30}\textbf{37.0} & \cellcolor{gray!30}\textbf{92.8} & \cellcolor{gray!30}\textbf{89.8} & \cellcolor{gray!30}\textbf{95.6} & \cellcolor{gray!30}\textbf{61.4} & \cellcolor{gray!30}\textbf{46.8} & \cellcolor{gray!30}\textbf{10.6} & \cellcolor{gray!30}\textbf{96.4} & \cellcolor{gray!30}\textbf{88.2} & \cellcolor{gray!30}\textbf{91.4} \\  
&\cellcolor{gray!20}\textit{avg. Imp} &\cellcolor{gray!20}\textit{\textbf{3.6$\uparrow$}} &\cellcolor{gray!20}\textit{\textbf{2.1$\uparrow$}} &\cellcolor{gray!20}\textit{\textbf{3.9$\uparrow$}} &\cellcolor{gray!20}\textit{\textbf{7.8$\uparrow$}} &\cellcolor{gray!20}\textit{\textbf{5.2$\uparrow$}} &\cellcolor{gray!20}\textit{\textbf{1.2$\uparrow$}} &\cellcolor{gray!20}\textit{\textbf{4.4$\uparrow$}} &\cellcolor{gray!20}\textit{\textbf{1.5$\uparrow$}} &\cellcolor{gray!20}\textit{\textbf{2.7$\uparrow$}} &\cellcolor{gray!20}\textit{\textbf{1.4$\uparrow$}} &\cellcolor{gray!20}\textit{\textbf{2.7$\uparrow$}} &\cellcolor{gray!20}\textit{\textbf{1.7$\uparrow$}} \\
& Upper Boundary & 39.2 & 47.0 & 37.6 & 95.0 & 95.2 & 96.8 & 65.0 & 48.6 & 12.6 & 97.2 & 88.4 & 92.4 \\ 
\bottomrule
\end{tabular}
}
\label{tab:comparison}
\end{table*}

\section{Experiments}
\subsection{Settings}
\textbf{Dataset.} We conduct experiments on two well-known benchmarks: MATHInstruct \cite{yue2023mammoth} and GLUE \cite{wang2018glue}. Specifically, MATHInstruct contains various mathematical tasks including fill-in-the-blank (FIBQ), multiple choice (MCQ), and programming questions (PQ). And GLUE is a collection of natural language understanding tasks such as single sentence classification (SSC), sentence pair classification (SPC), and natural language inference (NLI). To ensure balanced comparison across different tasks, we randomly sample 5,000 training examples and 500 test examples for each task.

\noindent\textbf{Baselines.} As shown in Table \ref{tab:models}, we design three model heterogeneous scenarios based on seven FMs, including Gemma \cite{team2024gemma}, Llama \cite{dubey2024llama}, SmolLM, and Yi \cite{young2024yi}, with parameter scales ranging from 1B to 8B. These models feature diverse architectural configurations with varying layer depths and hidden dimensions. To evaluate our proposed \textit{H$^2$Tune} comprehensively, we compare it against four representative baselines and an upper boundary. Specifically, \textit{LOCAL} trains models independently on each task without FL, serving as the lower performance bound. \textit{FLLM} represents the conventional federated setup where clients share identical foundation models and LoRA ranks, enabling direct parameter aggregation. \textit{HetLoRA} \cite{cho2024heterogeneous} allows clients to adopt different LoRA ranks while sharing the same base foundation models. \textit{FLTLA} aggregates only the tail layers across clients with different foundation models. As an upper boundary, we fine-tune local models on combined datasets of all clients without federation to indicate the maximum achievable performance.

\noindent\textbf{Implementation Details.} We conduct experiments using two GPU configurations with four L20 GPUs for models exceeding 5B parameters and four RTX 4090 GPUs for smaller models. The system involves 3 client nodes participating in the training process. We set the learning rates ranging from 2e-7 to 2e-3 with 5 communication rounds. Finally, we evaluate model performance through accuracy and quantify communication cost in MB and minutes.

\subsection{Main Result}
We evaluate \textit{H$^2$Tune} in both homogeneous and heterogeneous task scenarios. Table \ref{tab:comparison} presents the comprehensive results, demonstrating \textit{H$^2$Tune}'s superior performance across all scenarios. Our experimental results show that \textit{H$^2$Tune} achieves consistent improvements across all scenarios compared to baseline methods. In Scenario 1, \textit{H$^2$Tune} demonstrates performance gains of 4.3\% on average for MATHInstruct tasks and 3.7\% for GLUE tasks. Similar trends are observed in Scenario 2, where the model achieves improvements of 2.8\% and 3.0\% respectively. In Scenario 3, \textit{H$^2$Tune} shows average gains of 3.0\% on MATHInstruct and 3.3\% on GLUE tasks. While H$^2$Tune does not always reach the upper boundary performance, it maintains a reasonable gap while providing the significant advantage of using a single model for multiple tasks. Overall, these results clearly demonstrate that \textit{H$^2$Tune} can effectively improve model performance in both homogeneous and heterogeneous task scenarios.

\begin{table}[t]
\vspace{12pt}  
\centering
\caption{The influence of global ranks under heterogeneous and homogeneous scenarios.}
\resizebox{0.49\textwidth}{!}{
\begin{tabular}{lcccccc}
\toprule
\cellcolor{gray!30}Settings & \cellcolor{gray!30}Client 1 & \cellcolor{gray!30}Client 2 & \cellcolor{gray!30}Client 3 & \cellcolor{gray!30}Client 1 & \cellcolor{gray!30}Client 2 & \cellcolor{gray!30}Client 3 \\
\cellcolor{gray!30}Support max rank & \cellcolor{gray!30}16 & \cellcolor{gray!30}64 & \cellcolor{gray!30}64 & \cellcolor{gray!30}16 & \cellcolor{gray!30}64 & \cellcolor{gray!30}128 \\
\cellcolor{gray!30}Heterogeneous LLM & \cellcolor{gray!30}LLAMA-3B & \cellcolor{gray!30}GEMMA-2B & \cellcolor{gray!30}SmolLM-1.7B & \cellcolor{gray!30}LLAMA-3B & \cellcolor{gray!30}GEMMA-2B & \cellcolor{gray!30}SmolLM-1.7B \\ 
\midrule
\cellcolor{gray!20}Heterogeneous Task & \cellcolor{gray!20}FIBQ & \cellcolor{gray!20}MCQ &\cellcolor{gray!20}PQ & \cellcolor{gray!20}FIBQ & \cellcolor{gray!20}MCQ & \cellcolor{gray!20}PQ \\
\midrule
Global rank $r=64$ & 30.6 & 33.4 & 10.4 & - & - & - \\
Global rank $r=128$ & 31.0 & 33.5 & 10.7 & 31.6 & 34.8 & 17.2 \\
Global rank $r=192$ & 31.1 & 33.5 & 10.9 & 31.8 & 34.8 & 17.0 \\
\midrule
\cellcolor{gray!20}Heterogeneous Task & \cellcolor{gray!20}SSC & \cellcolor{gray!20}SPC & \cellcolor{gray!20}NLI & \cellcolor{gray!20}SSC & \cellcolor{gray!20}SPC &\cellcolor{gray!20}NLI \\
\midrule
Global rank $r=64$ & 95.4 & 86.8 & 84.8 & - & - & - \\
Global rank $r=128$ & 95.8 & 87.1 & 85.5 & 95.8 & 87.0 & 86.0 \\
Global rank $r=192$ & 96.4 & 87.5 & 85.8 & 96.2 & 87.6 &85.2 \\
\midrule
\cellcolor{gray!20}Homogeneous Task & \multicolumn{6}{c}{\cellcolor{gray!20}MCQ} \\
\midrule
Global rank $r=64$ & 27.0 & 33.2 & 17.3 & 27.5 & 33.7 & 18.2 \\
Global rank $r=128$ & 28.7 & 34.0 & 19.9 & 29.8 & 34.6 & 21.3 \\
Global rank $r=192$ & 28.6 & 34.2 & 18.5 & 29.4 & 35.0 & 20.1 \\
\bottomrule
\end{tabular}
}
\label{tab:influence_of_global_ranks}
\end{table}

\subsection{Ablation Study}
To investigate the effectiveness of proposed components in \textit{H$^2$Tune}, we conduct ablation experiments as shown in Table \ref{tab3}. Specifically, removing TSM causes the most performance drop, with average accuracy decreasing by 5.9\% on MATHInstruct and 19.1\% on GLUE, indicating its crucial role in capturing shared knowledge. Without ATKD, we observe an average performance reduction of 5.5\% on MATHInstruct and minimal impact on GLUE. And the removal of SM shows moderate performance degradation on MATHInstruct with a 1.7\% average drop, while maintaining comparable performance on GLUE tasks. This highlights that sparsification plays a task-dependent role in performance.

\begin{table}[t]
\centering
\caption{Results of ablation experiments, where TSM, ATKD, and SM represent Task-shared matrix, Alternating Task-Knowledge Disentanglement, and Sparsification Matrix, respectively.}
\label{tab3}
\resizebox{0.475\textwidth}{!}{
\begin{tabular}{l|ccc|ccc}
\toprule
 & FIBQ & MCQ  & PQ   & SSC  & SPC  & NLI  \\ 
\midrule
\cellcolor{gray!20}\textit{H$^2$Tune}   & \cellcolor{gray!20}\textbf{32.4} & \cellcolor{gray!20}\textbf{35.0} & \cellcolor{gray!20}\textbf{17.2} & \cellcolor{gray!20}\textbf{95.8} & \cellcolor{gray!20}\textbf{87.0}   & \cellcolor{gray!20}\textbf{86.0}   \\
\quad w/o TSM     & 28.8 & 27.2 & 10.8 & 90.4 & 66.6 & 54.4 \\
\quad w/o ATKD    & 29.0 & 27.8 & 11.2 & 95.4 & 86.2 & 86.8 \\
\quad w/o SM      & 30.2 & 29.4 & 19.8 & 95.8 & 86.4 & 88.2 \\
\bottomrule
\end{tabular}
}
\end{table}

\subsection{Communication Cost}
\begin{figure}[t]
\centering
\includegraphics[scale=0.25]{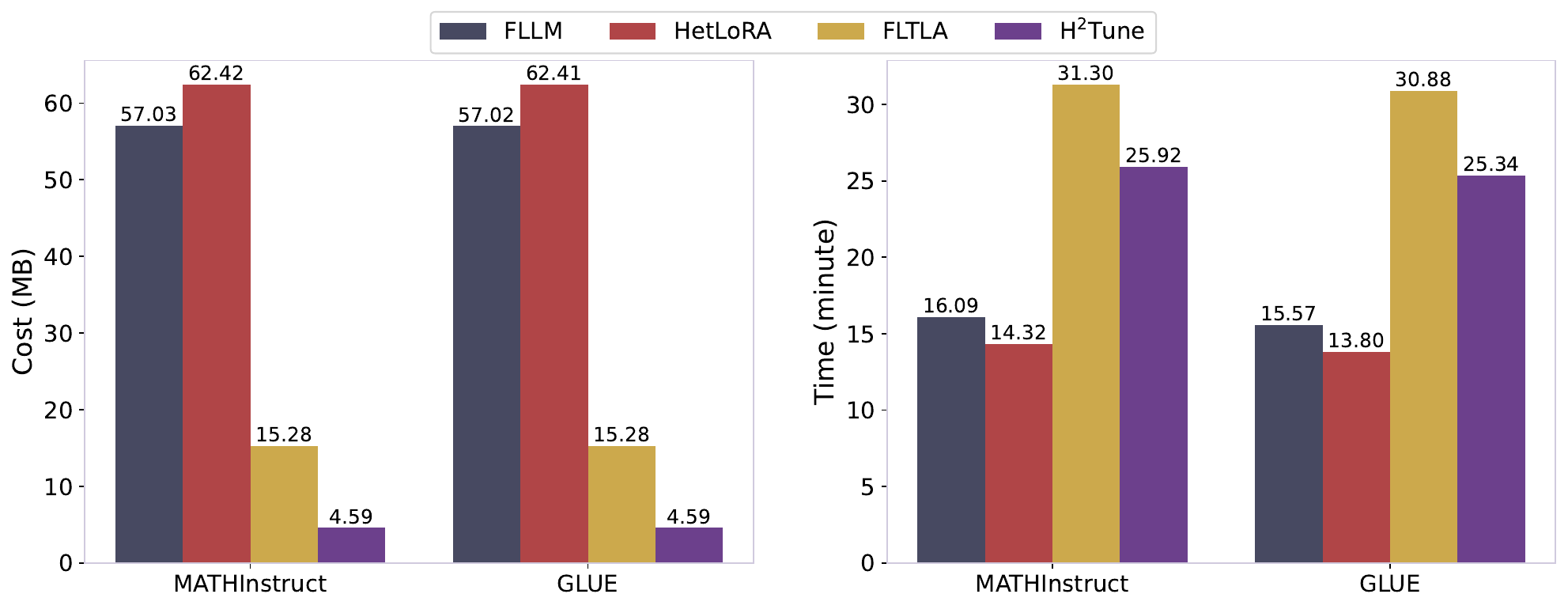}
\caption{Comparison of local and strategy communication costs.}
\label{fig: efficiency}
\vspace{8pt}  
\end{figure}

We evaluate the communication costs and training efficiency using MATHInstruct and GLUE benchmarks, as shown in Figure \ref{fig: efficiency}. While FLLM and HetLoRA require substantial communication overhead, our proposed H²Tune achieves efficient communication at only 4.59 MB, significantly outperforming FLTLA. In terms of training time, \textit{H$^2$Tune} shows similar computational overhead to FLTLA, which is moderately higher than FLLM and HetLoRA due to the task-knowledge disentanglement process. Notably, this slight increase in training time is well justified by the significant performance improvements demonstrated in our main results, while maintaining better communication efficiency compared with FLTLA.

\begin{figure}[t]
\centering
\includegraphics[scale=0.5]{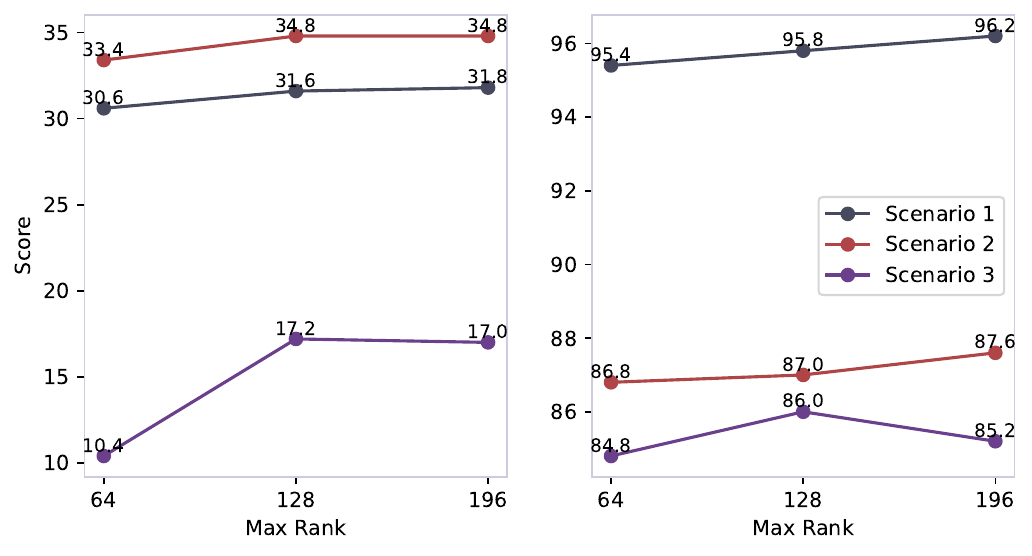}
\caption{Performance of heterogeneous models with different maximum ranks on MATHInstruct (left) and GLUE (right).}
\label{fig: difft}
\vspace{14pt}  
\end{figure}

\begin{table*}[t]
\caption{The influence of global layers under heterogeneous and homogeneous scenarios.}
\resizebox{\textwidth}{!}{
\begin{tabular}{llllllllllll}
\toprule
\multicolumn{3}{l}{\cellcolor{gray!30}Settings}                                     & \cellcolor{gray!30}Client 1 & \cellcolor{gray!30}Client 2 & \cellcolor{gray!30}Client 3    & \cellcolor{gray!30}Client 1 & \cellcolor{gray!30}Client 2 & \cellcolor{gray!30}Client 3    & \cellcolor{gray!30}Client 1 & \cellcolor{gray!30}Client 2   & \cellcolor{gray!30}Client 3   \\ 
\multicolumn{3}{l}{\cellcolor{gray!30}Heterogeneous LLM}                            & \cellcolor{gray!30}LLaMA-3B & \cellcolor{gray!30}GEMMA-2B & \cellcolor{gray!30}SmolLM-1.7B & \cellcolor{gray!30}LLaMA-3B & \cellcolor{gray!30}GEMMA-2B & \cellcolor{gray!30}SmolLM-1.7B & \cellcolor{gray!30}LLaMA-3B & \cellcolor{gray!30}LLaMA-3.2B & \cellcolor{gray!30}LLaMA-3.2B \\ \hline
\multicolumn{3}{l}{\cellcolor{gray!20}Heterogeneous Task}                           & \cellcolor{gray!20}FIBQ     & \cellcolor{gray!20}MCQ      & \cellcolor{gray!20}PQ          & \cellcolor{gray!20}SSC      & \cellcolor{gray!20}SPC      & \cellcolor{gray!20}NLI         & \cellcolor{gray!20}FIBQ     & \cellcolor{gray!20}MCQ        & \cellcolor{gray!20}PQ         \\ \hline
\multicolumn{3}{l}{\cellcolor{gray!10}Real layer}                                   & \cellcolor{gray!10}28       & \cellcolor{gray!10}18       & \cellcolor{gray!10}24          & \cellcolor{gray!10}28       & \cellcolor{gray!10}18       & \cellcolor{gray!10}24          & \cellcolor{gray!10}32       & \cellcolor{gray!10}28         & \cellcolor{gray!10}16         \\ \hline
\multirow{6}{*}{Global layer}     & \multicolumn{2}{l}{L=18}     & 30.4     & 29.2     & 15.4        & 49.7     & 82.3     & 78.9        & 60.4     & 40.6       & 5.1        \\
                                  & \multicolumn{2}{l}{L=24}     & 32.4     & 29.2     & 16.6        & 51.3     & 84.7     & 82.1        & 62.8     & 42.9       & 7.5        \\
                                  & \multicolumn{2}{l}{L=28}     & 34.8     &31.6     & 17.2        & 52.8     & 87.4     & 83.7        & 63.9     & 44.1       & 8.4        \\
                                  & \multicolumn{2}{l}{L=32}     & 31.4     & 31.6     & 17.2        & 53.0     & 88.4     & 85.2        & 64.1     & 46.8       & 10.6       \\
                                  & \multicolumn{2}{l}{L=40}     & 31.6     & 31.8     & 20.8        & 52.2     &86.9     & 84.8        &63.2     & 45.5       & 9.2        \\
                                  & \multicolumn{2}{l}{L=48}     & 29.6     & 32.4     & 15.4        & 47.9     & 85.6     & 84.0        & 59.4     & 45.3       & 8.8        \\ \hline
\multicolumn{3}{l}{\cellcolor{gray!20}Homogeneous Task}                             & \multicolumn{9}{c}{\cellcolor{gray!20}MCQ}                                                                                    \\ \hline
\multicolumn{3}{l}{\cellcolor{gray!10}Real layer}                                   & \cellcolor{gray!10}28       & \cellcolor{gray!10}18       & \cellcolor{gray!10}24          & \cellcolor{gray!10}28       & \cellcolor{gray!10}18       & \cellcolor{gray!10}24          & \cellcolor{gray!10}32       & \cellcolor{gray!10}28         & \cellcolor{gray!10}16         \\ \hline
\multicolumn{2}{l}{\multirow{6}{*}{Global layer}} & L=18         & 26.8     & 31.2     & 18.7        & 38.4     & 29.7     & 24.7        & 33.5     & 41.8       & 35.3       \\
\multicolumn{2}{l}{}                              & L=24         & 27.7     & 33.1     & 20.0        & 40.7     & 32.5     & 24.3        & 34.2     & 42.9       & 35.6       \\
\multicolumn{2}{l}{}                              & L=28         & 28.4     & 33.8     & 19.8        & 41,3     & 32.9     & 24.7        & 35.1     & 41.8       & 36.0       \\
\multicolumn{2}{l}{}                              & L=32         & 29.2     & 34.4     & 20.6        & 42.2     & 33.2     & 25.2        & 36.0     & 43.6       & 37.0       \\
\multicolumn{2}{l}{}                              & L=40         & 28.6     & 33.7     & 20.3        & 41.8     & 32.4     & 24.5        & 35.5     & 40.7       & 35.8       \\
\multicolumn{2}{l}{}                              & L=48         & 27.9     & 33.9     & 19.2        & 42.1     & 32.0     & 24.1        & 35.8     & 42,9       & 36.9    \\
\bottomrule
\end{tabular}}
\label{tab:global layers}
\end{table*}

\subsection{Influence of Global Ranks}
\noindent\textbf{Homogeneous task scenario.} Table \ref{tab:influence_of_global_ranks} presents the impact of increasing global rank on model performance in homogeneous task scenario where all clients perform MCQ tasks. Results indicate that increasing the global rank from 64 to 128 yields consistent performance improvements across all clients. Further increasing the global rank to 192 produces minimal additional gains in some cases, suggesting that the benefits of higher rank representations plateau after a certain threshold. Notably, the configuration with support max rank of 16/64/128 consistently outperforms the 16/64/64 configuration across all global ranks, with client 3 showing the most significant improvement when given a higher maximum rank capacity.

\noindent\textbf{Heterogeneous task scenario.} For heterogeneous task settings, we observe nuanced performance patterns across different client configurations. In the first task set, lifting the global rank from 64 to 128 yields gains of 0.4, 0.1, and 0.3 percentage points for clients 1, 2, and 3, respectively. For the second task set involving classification tasks, performance improvements are more pronounced as the global rank increases, with consistent gains across SSC, SPC, and NLI tasks. 

\noindent\textbf{Different model heterogeneity scenarios.} Figure \ref{fig: difft} shows the results of Gemma-2B, Llama-3B, and SmolLM-1.7B at different maximum ranks using the \textit{H$^2$Tune} strategy, where the maximum ranks for the three clients varying from 64 to 196. The results demonstrate that increasing the maximum rank from 64 to 128 improves performance. However, further increasing to 196 yields marginal gains, indicating that an appropriate rank setting helps balance between feature expressiveness and potential noise introduction.

\begin{figure}[t]
\centering
\includegraphics[scale=0.31]{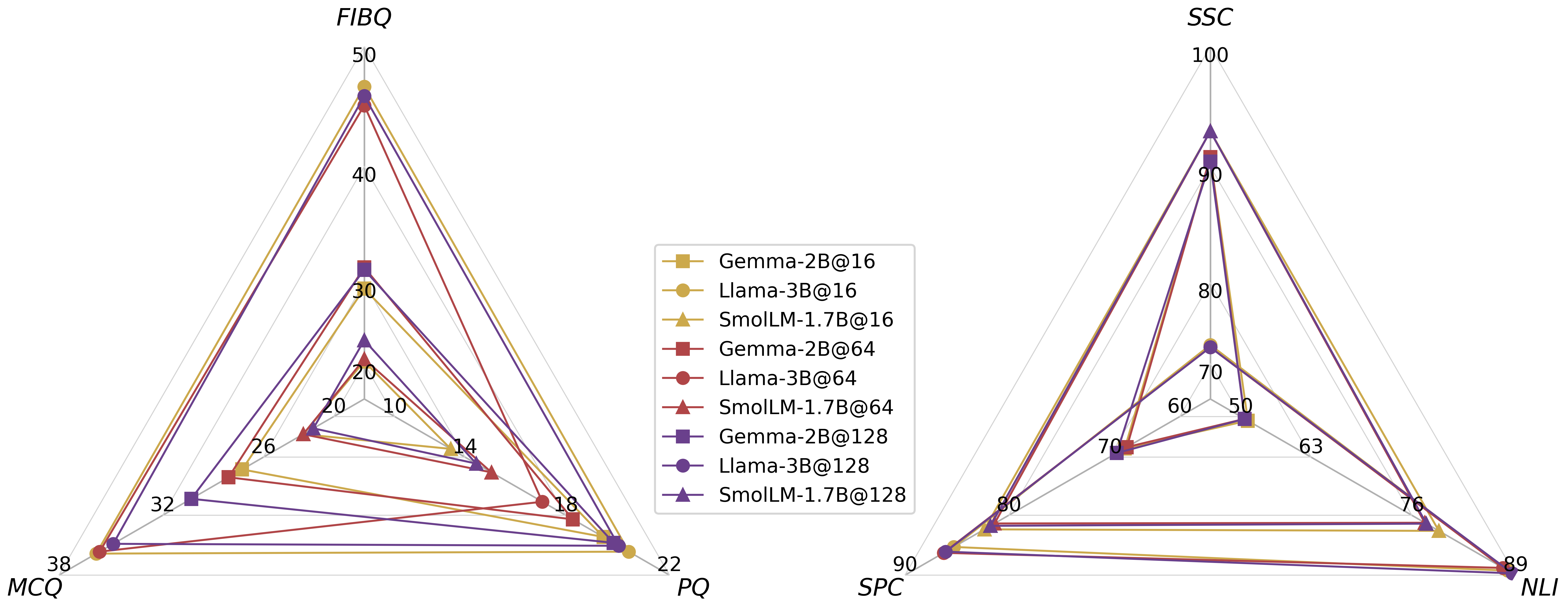}
\caption{Performance of homogeneous models with different ranks. $@r$ denotes the resource-constrained rank for client updates.}
\label{fig: lorar}
\vspace{14pt}  
\end{figure}

\subsection{Influence of Global Layers}
\noindent\textbf{Homogeneous task scenario.} Table \ref{tab:global layers} presents the impact of global layer count on model performance in homogeneous task scenario. As global layer count increases from 18 to 32, LLaMA-3B, GEMMA-2B, and SmolLM-1.7B show accuracy gains of 2.4\%, 3.2\%, and 1.9\% in scenario 1. Similar trends appear in Scenarios 2 and 3. However, further increasing to 40 or 48 layers yields diminishing returns or even slight performance decreases.

\noindent\textbf{Heterogeneous task scenario.} For heterogeneous tasks, we observe more varied responses to global layer count increases. For example, FIBQ performance peaks at $L=28$ with 34.8\% accuracy, while MCQ and PQ achieve optimal results at $L=48$ and $L=40$ with 32.4\% and 20.8\% accuracy in scenario 1. The results indicate heterogeneous scenarios generally require higher global layer, with most optimal performance observed between $L=28$ and $L=32$.

\begin{table}[t]
    \centering
    \caption{Hyper-parameter analysis results for $\beta_k$.}
    \begin{tabular}{lccccc}
        \toprule
        Model & $\beta_k$ & Task & MATHInstruct & Task & GLUE \\
        \midrule
        \multirow{3}{*}{LLaMA-3B} & 0.125 & \multirow{3}{*}{FIBQ} & 31.6 & \multirow{3}{*}{SSC} & 95.3 \\
         & 0.500 &  & 32.4 &  & 95.8 \\
         & 1.000 &  & 32.2 &  & 95.4 \\
        \midrule
        \multirow{3}{*}{GEMMA-2B} & 0.125 & \multirow{3}{*}{MCQ} & 32.4 & \multirow{3}{*}{SPC} & 86.2 \\
         & 0.500 &  & 34.8 &  & 87.0 \\
         & 1.000 &  & 35.0 &  & 86.8 \\
        \midrule
        \multirow{3}{*}{SmolLM-1.7B} & 0.125 & \multirow{3}{*}{PQ} & 11.8 & \multirow{3}{*}{NLI} & 86.0 \\
         & 0.500 &  & 17.0 &  & 85.1 \\
         & 1.000 &  & 17.2 &  & 85.9 \\
        \bottomrule
    \end{tabular}
    \label{tab: hyper-parameter analysis}
\end{table}
\subsection{Hyper-parameter Analysis}
We investigate how the local parsity ratio $\beta_k$ affects model performance in both homogeneous and heterogeneous model settings.

\noindent\textbf{Homogeneous models with different local ranks.} Figure \ref{fig: lorar} illustrates the performance of Gemma-2B, Llama-3B, and SmolLM-1.7B across different ranks in a homogeneous model setting, where all clients use identical models and ranks. The performance is visualized as radar charts, where a larger triangle area indicates better overall performance, and the angle bias reflects performance imbalance across tasks. We observe that increasing rank values consistently leads to better performance, suggesting that larger LoRA parameters can capture more task-relevant information.

\noindent\textbf{Heterogeneous models with different local ranks.} As shown in Table \ref{tab: hyper-parameter analysis}, we observe that increasing sparsity ratio $\beta_k$ generally improves performance across models and tasks, with optimal values varying by task type. These results indicate that mathematical reasoning tasks benefit from higher sparsity ratios, while language understanding tasks often perform optimally at moderate values around $\beta_k=0.500$, suggesting task-specific adaptation of sparsity is beneficial for heterogeneous model deployments.

\section{Conclusion}
This paper explores an under-explored direction: federated fine-tuning with heterogeneous models, tasks, and resources. We propose triple matrix decomposition and layer relation matrices to handle heterogeneous matrix fusion. Our alternating optimization separates shared and private knowledge to mitigate multi-task interference, enabling cross-client knowledge transfer. Extensive experiments show our method improves accuracy by up to 15.4\% over baselines.






\bibliography{ecai-sample-and-instructions}

\clearpage

\appendix
                         
\section{Algorithm Pseudocode}
\begin{algorithm}[h]
\caption{$\mathit{H^2Tune}$}
\begin{algorithmic} \label{algorithm flowchart}
\STATE{\textbf{Input:} communication round $T$; learning rate: $\eta$ and $\eta^{\prime}$; epoch: $\tau$; training dataset $(x_i,y_i)\in D_k$ of client $k$.}
	 \FOR{$t=0, 1, \dots, T-1$}
        \STATE{parties receive global task-shared matrix $\mathbf{R_g^t}$;}
         \FOR{party $k=1,...,K$ in parallel} 
             \FOR{$j=1,\dots, \tau$}
                 \FOR{$(x_i,y_i)\in D_k$}
                     \STATE{\textbf{Freeze} task-specific matrices $\mathbf{A_{k}}$ and $\mathbf{B_{k}}$;}
                     \STATE{update $\mathbf{\Phi^{t,j}_k}\mathbf{R^{t,j}_{k}},  = \mathbf{\Phi^{t,j}_k}\mathbf{R^{t,j-1}_{k}} - \eta' \nabla_{\mathbf{R}} \mathcal{L}_{share}(x_i,y_i)$;}
                     \STATE{update $\mathbf{\Omega^{t,j}_{k}},  = \Omega^{t,j-1}_{k} - \eta' \nabla_{\mathbf{\Omega}} \mathcal{L}_{share}(x_i,y_i)$;}
                     \STATE{\textbf{Freeze} task-shared matrix $\mathbf{\Phi_k}\mathbf{R_{k}}$ and relation matrix $\Omega_k$;}
                     \STATE{update $\mathbf{A^{t,j}_k} = \mathbf{A^{t,j}_k} - \eta \nabla_{\mathbf{A}} \mathcal{L}_{specific}(x_i,y_i)$;}
                     \STATE{update $\mathbf{B^{t,j}_k} = \mathbf{B^{t,j}_k} - \eta \nabla_{\mathbf{B}} \mathcal{L}_{specific}(x_i,y_i)$;}
                \ENDFOR
              \ENDFOR
              \STATE{upload task-shared matrix $\mathbf{R_k^t}$ to server;}
            \ENDFOR
          \STATE{server aggregates $\mathbf{R_k^t}$ from party $k =1, \cdots, K$ by $\mathbf{R_g^t}=\frac{1}{K}\sum^{K}_{k=1}\mathbf{R_k^t}$;}
      \ENDFOR
\end{algorithmic} 
\end{algorithm}

\section{Convergence Analysis}
\label{sec:convergence analysis}
We begin with introducing the convergence metric $\|\mathcal{G}_t^j\|$ to measure convergence properties of our algorithms. We define the generalized gradient at the $j$-th iteration of $t$-th round:
\begin{equation*}
    \mathcal{G}_t^j = \frac{1}{\eta'} \left(\mathbf{R}_t^j - \mathbf{R}_t^{j+1}\right).
\end{equation*}
When $j\leq \tau-2$, the $j$-th iteration corresponds to the gradient descent step, we have $\| \mathcal{G}_t^j\| = \| \nabla F(\mathbf{R}_t^j)\|$ which is a common convergence metric used in \cite{ghadimi2018approximation,ji2021bilevel}. When $j=\tau-1$, the $j$-th iteration corresponds to the proximal step, where $\| \mathcal{G}_t^j\|$ was also used as an important metric (\cite{huang2022enhanced}). To simplify the convergence analysis, we can rewrite Algorithm \ref{algorithm flowchart} as Algorithm \ref{algorithm bilevel}, and present the convergence analysis under Algorithm \ref{algorithm bilevel}.

\begin{algorithm}[h]
    \caption{Proximity Version of $\mathit{H^2Tune}$}
    \begin{algorithmic}\label{algorithm bilevel}
        \STATE{\textbf{Input:} communication round $T$; learning rate: $\eta$ and $\eta^{\prime}$; epoch: $\tau$.}
        \FOR{$t=0, 1, \dots, T-1$}
            \STATE{Let $\mathbf{H}_t^0=\mathbf{H}_{t-1}^\tau$ and $\mathbf{R}_t^0=\mathbf{R}_{t-1}^\tau$;}
            \FOR{$j= 1,2, \dots, \tau-1$}
                \STATE{update $\mathbf{R}_{t}^j=\mathbf{R}_{t}^{j-1} - \eta^\prime \nabla_{\mathbf{R}} \mathcal{L}(\mathbf{R}_{t}^{j-1}, \mathbf{H}_{t-1}^\tau)$;}
            \ENDFOR
        \STATE{update $\mathbf{R}_{t}^\tau=\arg\min_{\mathbf{R}}\{ \langle \frac{\partial \mathcal{L}(\mathbf{R}_t^{\tau-1}, \mathbf{H}_{t-1}^\tau)}{\partial \mathbf{R}_{t}^{\tau-1}}, \mathbf{R} \rangle + h(\mathbf{R}) + \frac{1}{\eta^\prime}\|\mathbf{R} - \mathbf{R}_{t}^{\tau-1} \|^2 \}$}
        \FOR{$j= 1,2, \dots, \tau$}
        \STATE{update $\mathbf{H}_{t}^j=\mathbf{H}_{t}^{j-1} - \eta \nabla_{\mathbf{H}} \mathcal{G}(\mathbf{R}_{t}^{\tau}, \mathbf{H}_{t}^{j-1})$;}
        \ENDFOR
        \ENDFOR
    \end{algorithmic}
\end{algorithm}
In Algorithm \ref{algorithm bilevel}, $\mathbf{R}_{t}^\tau=\arg\min_{\mathbf{R}}\{ \langle \frac{\partial \mathcal{L}(\mathbf{R}_t^{\tau-1}, \mathbf{H}_{t-1}^\tau)}{\partial \mathbf{R}_{t}^{\tau-1}}, \mathbf{R} \rangle \allowbreak + h(\mathbf{R}) + \frac{1}{\eta^\prime}\|\mathbf{R} - \mathbf{R}_{t}^{\tau-1} \|^2 \}$ is equivalent to doing the last step in Algorithm \ref{algorithm flowchart} first, and then assigns a personalized parameter $\mathbf{R}_k$ for each client.

\subsection{Proof Sketch}
\noindent \textbf{Proof Sketch.} To explore the essential insights, we first bound the difference between the local parameter and the local optimal parameter at the $t$-th round (i.e. tracking error $\|\mathbf{H}_{t}^{j-1} - \mathbf{H}^*(\mathbf{R}_t^0) \|$). Then, the upper bound of the difference between the approximate gradient and the exact gradient (i.e. $\Big\| \frac{\partial \mathcal{L}(\mathbf{R}_t^j, \mathbf{H}_t^\tau)}{\partial \mathbf{R}_t^j} - \nabla F(\mathbf{R}_t^j) \Big\|$) is provided by a proof technology called virtual updates (\cite{yang2022fastslowmo}). Lastly, we make use of the property of the proximal operator in our bi-level optimization problem to deal with the proximal step in our algorithm. 

\subsection{Proof Details}
To provide the convergence analysis of our algorithm, we need to give some auxiliary lemmas first.
Firstly, with $L$-smoothness of ${G}_0(\mathbf{R}, \mathbf{H})$, we could obtain the strongly-convexity property of ${G}(\mathbf{R}, \mathbf{H})$.
\begin{lemma}
Under Assumption \ref{ass 2}, suppose $\alpha := -L + \mathcal{V} > 0$, ${G}(\mathbf{R}, \mathbf{H})$ is $\alpha$-strongly convex w.r.t $\mathbf{H}$.
\label{lemma:1}
\end{lemma}
The proof of this lemma is trivial. Thus, we omit it.

\begin{lemma}[Lemma 2.2 in \cite{ghadimi2018approximation}]
Under Assumptions \ref{ass 1}, \ref{ass 2}, and \ref{ass 3}, $F(\mathbf{R})$ is $L_0$-smooth, where $L_0$ is given by
    \begin{equation}
    \small
         L_0 := L_2 + \frac{2 L_2^2+L_1^2L_3}{\alpha} + \frac{L_1 L_2 L_3 + L_1 L_2 L_4 + L_2^3}{\alpha^2} + \frac{L_1 L_2^2 L_4}{\alpha^3}.
    \end{equation}
\label{lemma:2}
\end{lemma}
Tracking error $\|\mathbf{H}_t^{j-1} - \mathbf{H}^{\ast}(\mathbf{R}_t^0)\|$ is an important component in our convergence analysis. To give an upper bound on the tracking error, we utilized Lemma 9 in \cite{ji2021bilevel}.
\begin{lemma}[Lemma 9 in \cite{ji2021bilevel}]
Under Assumptions \ref{ass 1} and \ref{ass 2}, with step size $\eta$ to be $\frac{2}{L_2 + \alpha}$, we have
\begin{equation} \label{eq:tracking_error}
\|\mathbf{H}_t^{(j-1)} - \mathbf{H}^{\ast}\big(\mathbf{R}_t^0\big)\| 
\leq \left(\frac{L_2 - \alpha}{L_2 + \alpha}\right)^{(j-1)} 
\|\mathbf{H}_t^{0} - \mathbf{H}^{\ast}\big(\mathbf{R}_t^{0}\big)\|.
\end{equation}
\label{lemma:3}
\end{lemma}
With the help of the above lemmas, we could give the estimation property of the $\frac{\partial \mathcal{L}(\mathbf{R}_t^0, \mathbf{H}_t^\tau)}{\partial \mathbf{R}_t^0}$ approximating $\nabla F(\mathbf{R}_t^0)$. The result is presented in the following Proposition \ref{prop:1}.
\begin{proposition}
Under Assumptions \ref{ass 1}-\ref{ass 5}, choose step size $\eta$ to be $\frac{2}{L_2 + \alpha}$, we have
\begin{equation}
\small
\begin{aligned}
&\left\| \frac{\partial L(\mathbf{R}_t^0, \mathbf{H}_t^\tau)}{\partial \mathbf{R}_t^0} - \nabla F(\mathbf{R}_t^0) \right\| 
\leq \left(L_2 + \frac{L_2^2}{\alpha}\right) \left[\left(\frac{L_2 - \alpha}{L_2 + \alpha}\right)^\tau \cdot \Delta\right] \\
&\quad + L_1 \Bigg[
\frac{L_2 \left(1 - \frac{2}{L_2 + \alpha} \cdot \alpha \right)^\tau}{\alpha} 
+ \frac{2}{L_2 + \alpha} \left(\frac{L_2 L_4}{\alpha} + L_3 \right) \cdot \Delta \\
&\quad \quad 
\frac{\left(1 - \frac{2}{L_2 + \alpha} \cdot \alpha \right)^\tau}{1 - \frac{2}{L_2 + \alpha} \cdot \alpha - \frac{L_2 - \alpha}{L_2 + \alpha}}
\Bigg].
\label{eq:2}
\end{aligned}
\end{equation}
\label{prop:1}
\end{proposition}

\noindent\textbf{Proof.} Using:
\begin{equation}
\begin{aligned}
\nabla F(\mathbf{R}_t^0) &= \nabla_{\mathbf{R}} \mathcal{L}(\mathbf{R}_t^0, \mathbf{H}^{\ast}(\mathbf{R}_t^0)) \\
&+ \frac{\partial \mathbf{H}^{\ast}(\mathbf{R}_t^0)}{\partial \mathbf{R}_t^0} \nabla_{\mathbf{H}} \mathcal{L}(\mathbf{R}_t^0, \mathbf{H}^{\ast}(\mathbf{R}_t^0)),
\end{aligned}
\label{eq:3}
\end{equation}
and:
\begin{equation}
\frac{\partial \mathcal{L}(\mathbf{R}_t^0, \mathbf{H}_t^\tau)}{\partial \mathbf{R}_t^0} = \nabla_{\mathbf{R}} \mathcal{L}(\mathbf{R}_t^0, \mathbf{H}_t^\tau) 
+ \frac{\partial \mathbf{H}_t^\tau}{\partial \mathbf{R}_t^0} \nabla_{\mathbf{H}} \mathcal{L}(\mathbf{R}_t^0, \mathbf{H}_t^\tau),
\label{eq:4}
\end{equation}
we have:
\begin{footnotesize}
\begin{equation}
\begin{aligned}
&\Big\| \frac{\partial \mathcal{L}(\mathbf{R}_t^0, \mathbf{H}_t^\tau)}{\partial \mathbf{R}_t^0} - \nabla F(\mathbf{R}_t^0) \Big\| 
\leq L_2 \left\|\mathbf{H}_t^\tau - \mathbf{H}^*(\mathbf{R}_t^0)\right\| + L_1 \Big\| \frac{\partial \mathbf{H}_t^\tau}{\partial \mathbf{R}_t^0} \\
&- \frac{\partial \mathbf{H}^{\ast}(\mathbf{R}_t^0)}{\partial \mathbf{R}_t^0} \Big\| + L_2 \Big\|\frac{\partial \mathbf{H}^{\ast}(\mathbf{R}_t^0)}{\partial \mathbf{R}_t^0} \Big\| \cdot \|\mathbf{H}_t^\tau - \mathbf{H}^{\ast}(\mathbf{R}_t^0)\|.
\end{aligned}
\label{eq:5}
\end{equation}
\end{footnotesize}
Now we first bound:
\(\left\| \frac{\partial \mathbf{H}_t^\tau}{\partial \mathbf{R}_t^0} - \frac{\partial \mathbf{H}^{\ast}(\mathbf{R}_t^0)}{\partial \mathbf{R}_t^0} \right\|\).
Recall the update method:
\begin{equation}
\mathbf{H}_t^j = \mathbf{H}_t^{j-1} - \eta \nabla_{\mathbf{H}} {G}(\mathbf{R}_t^0, \mathbf{H}_t^{j-1}),
\label{eq:6}
\end{equation}
and we apply the chain rule on it, we have:
\begin{equation}
\begin{aligned}
\frac{\partial \mathbf{H}_t^j}{\partial \mathbf{R}_t^0} &= \frac{\partial \mathbf{H}_t^{j-1}}{\partial \mathbf{R}_t^0} 
- \eta \Big(\nabla_{\mathbf{R}} \nabla_{\mathbf{H}} {G}(\mathbf{R}_t^0, \mathbf{H}_t^{j-1}) \\
&+ \frac{\partial \mathbf{H}_t^{j-1}}{\partial \mathbf{R}_t^0} \nabla_{\mathbf{H}}^2 {G}(\mathbf{R}_t^0, \mathbf{H}_t^{j-1})\Big).
\end{aligned}
\label{eq:7}
\end{equation}
Since \(\mathbf{H}^*(\mathbf{R}_t^0)\) is the optimal solution of \({G}(\mathbf{R}_t^0, \mathbf{H})\), we have $\nabla_{\mathbf{H}} {G}(\mathbf{R}_t^0, \mathbf{H}^*(\mathbf{R}_t^0)) = 0.$
Then, using the chain rule, we have:
\begin{equation}
\small
\nabla_{\mathbf{R}} \nabla_{\mathbf{H}} {G}\big(\mathbf{R}_t^0, \mathbf{H}^{\ast}(\mathbf{R}_t^0)\big) 
+ \frac{\partial \mathbf{H}^*(\mathbf{R}_t^0)}{\partial \mathbf{R}_t^0} \nabla_{\mathbf{H}}^2 {G}(\mathbf{R}_t^0, \mathbf{H}^{\ast}\big(\mathbf{R}_t^0)\big) = 0.
\label{eq:8}
\end{equation}
Combining \eqref{eq:7} and \eqref{eq:8}, the following equation holds:
\begin{equation}
\small
\begin{aligned}
&\frac{\partial \mathbf{H}_t^j}{\partial \mathbf{R}_t^0} - \frac{\partial \mathbf{H}^{\ast}(\mathbf{R}_t^0)}{\partial \mathbf{R}_t^0} 
= \frac{\partial \mathbf{H}_t^{j-1}}{\partial \mathbf{R}_t^0} - \frac{\partial \mathbf{H}^{\ast}(\mathbf{R}_t^0)}{\partial \mathbf{R}_t^0} \\
&\quad- \eta \bigg( \nabla_{\mathbf{R}} \nabla_{\mathbf{H}} {G}(\mathbf{R}_t^0, \mathbf{H}_t^{j-1}) 
- \nabla_{\mathbf{R}} \nabla_{\mathbf{H}} {G}(\mathbf{R}_t^0, \mathbf{H}^{\ast}(\mathbf{R}_t^0)) \bigg) \\
&\quad - \eta \bigg( \frac{\partial \mathbf{H}_t^{j-1}}{\partial \mathbf{R}_t^0} 
- \frac{\partial \mathbf{H}^{\ast}(\mathbf{R}_t^0)}{\partial \mathbf{R}_t^0} \bigg) 
\nabla_{\mathbf{H}}^2 {G}(\mathbf{R}_t^0, \mathbf{H}_t^{j-1}) \\
&\quad + \eta \frac{\partial \mathbf{H}^{\ast}(\mathbf{R}_t^0)}{\partial \mathbf{R}_t^0} 
\bigg( \nabla_{\mathbf{H}}^2 {G}(\mathbf{R}_t^0, \mathbf{H}^{\ast}(\mathbf{R}_t^0)) 
- \nabla_{\mathbf{H}}^2 {G}(\mathbf{R}_t^0, \mathbf{H}_t^{j-1}) \bigg).
\end{aligned}
\label{eq:9}
\end{equation}
Combining \eqref{eq:8} and Assumption~\ref{ass 2} yields:
\begin{equation}
\left\| \frac{\partial \mathbf{H}^{\ast}(\mathbf{R}_t^0)}{\partial \mathbf{R}_t^0} \right\| \leq \frac{L_2}{\alpha}.
\label{eq:10}
\end{equation}
With the help of Lemma~\ref{lemma:2}, \eqref{eq:9} and \eqref{eq:10}, the following bound holds:
\begin{equation}
\small
\begin{aligned}
\left\| \frac{\partial \mathbf{H}_t^j}{\partial \mathbf{R}_t^0} - \frac{\partial \mathbf{H}^{\ast}(\mathbf{R}_t^0)}{\partial \mathbf{R}_t^0} \right\| 
&\leq (1 - \eta \alpha) \left\| \frac{\partial \mathbf{H}_t^{j-1}}{\partial \mathbf{R}_t^0} - \frac{\partial \mathbf{H}^{\ast}(\mathbf{R}_t^0)}{\partial \mathbf{R}_t^0} \right\| \\
&+ \eta \left(\frac{L_2L_4}{\alpha} + L_3 \right) \left\| \mathbf{H}_t^{j-1} - \mathbf{H}^{\ast}(\mathbf{R}_t^0) \right\|.
\end{aligned}
\label{eq:11}
\end{equation}
For the choice of \(\eta = \frac{2}{L_2 + \alpha}\), Lemma~\ref{lemma:3} holds. With Lemma~\ref{lemma:3} and Assumption~\ref{ass 4}, we have:
\begin{equation}
\begin{aligned}
\|\mathbf{H}_t^{j-1} - \mathbf{H}^{\ast}(\mathbf{R}_t^0)\| &\leq \left(\frac{L_2 - \alpha}{L_2 + \alpha}\right)^{j-1} 
\|\mathbf{H}_t^0 - \mathbf{H}^{\ast}(\mathbf{R}_t^0)\| \\
&\leq \left(\frac{L_2 - \alpha}{L_2 + \alpha}\right)^{j-1} \cdot \Delta.
\label{eq:12}
\end{aligned}
\end{equation}
Telescoping  \eqref{eq:11} over \(j\) from \(0\) to \(\tau\) and combining \eqref{eq:12} yields:
\begin{equation}
\begin{aligned}
\left\| \frac{\partial \mathbf{H}_t^\tau}{\partial \mathbf{R}_t^0} - \frac{\partial \mathbf{H}^{\ast}(\mathbf{R}_t^0)}{\partial \mathbf{R}_t^0} \right\|
&\leq \frac{L_2 (1 - \eta \alpha)^\tau}{\alpha} + \eta \left(\frac{L_2 L_4}{\alpha} + L_3\right) \\
&\Delta \cdot \frac{(1 - \eta \alpha)^\tau}{1 - \eta \alpha - \frac{L_2 - \alpha}{L_2 + \alpha}}.
\end{aligned}
\label{eq:13}
\end{equation}
Plugging \eqref{eq:10} \eqref{eq:12} \eqref{eq:13} into \eqref{eq:5} yields:
\begin{equation}
\small
\begin{aligned}
&\Big\| \frac{\partial L(\mathbf{R}_t^0, \mathbf{H}_t^\tau)}{\partial \mathbf{R}_t^0} - \nabla F(\mathbf{R}_t^0) \Big\|
\leq \left(L_2 + \frac{L_2^2}{\alpha}\right) \left[\left(\frac{L_2 - \alpha}{L_2 + \alpha}\right)^\tau \cdot \Delta\right] \\
&+ L_1 \Bigg[\frac{L_2 (1 - \frac{2}{L_2 + \alpha} \cdot \alpha)^\tau}{\alpha} 
+ \frac{2}{L_2 + \alpha} \left(\frac{L_2 L_4}{\alpha} + L_3 \right) \Delta \\
&\cdot \frac{\left(1 - \frac{2}{L_2 + \alpha} \cdot \alpha \right)^\tau}{1 - \frac{2}{L_2 + \alpha} \cdot \alpha - \frac{L_2 - \alpha}{L_2 + \alpha}}
\Bigg].
\end{aligned}
\label{eq:14}
\end{equation}
Hence, we complete the proof of Proposition~\ref{prop:1}.

To give an upper bound on $\Big\| \frac{\partial \mathcal{L}(\mathbf{R}_t^j, \mathbf{H}_t^\tau)}{\partial \mathbf{R}_t^j} - \nabla F(\mathbf{R}_t^j) \Big\|$, we make use of a technology called virtual updates (\cite{yang2022fastslowmo}). Specifically, we introduce a virtual parameter $\mathbf{H}_{t,j}^\tau$ obtained by $\tau$ updates when $\mathbf{R}_t^j$ is given. Using Proposition~\ref{prop:1}, 
We can bound $\Big\| \frac{\partial \mathcal{L}(\mathbf{R}_t^j, \mathbf{H}_{t,j}^\tau)}{\partial \mathbf{R}_t^j} - \nabla F(\mathbf{R}_t^j) \Big\|$.
Now, we aim to bound $\Big\|\frac{\partial \mathcal{L}(\mathbf{R}_t^j, \mathbf{H}_t^\tau)}{\partial \mathbf{R}_t^j} - \nabla F(\mathbf{R}_t^j)\Big\|$.
The result is shown in Proposition \ref{prop:2}.
\begin{proposition}
Following the conditions in Proposition~\ref{prop:1}, we have:
\begin{equation}
\small
\begin{aligned}
&\Big\| \frac{\partial \mathcal{L}(\mathbf{R}_t^j, \mathbf{H}_t^\tau)}{\partial \mathbf{R}_t^j} - \nabla F(\mathbf{R}_t^j) \Big\|
\leq \Big(L_2 + \frac{L_2^2}{\alpha}\Big) \Big[\left(\frac{L_2 - \alpha}{L_2 + \alpha}\right)^\tau \cdot \Delta \Big] \\
&\quad + L_1 \Bigg[\frac{L_2 \Big(1 - \frac{2}{L_2 + \alpha}  \Big)^\tau}{\alpha} 
+ \frac{2}{L_2 + \alpha} \Big(\frac{L_2 L_4}{\alpha} + L_3\Big) \Delta \\
&\quad \quad \frac{\Big(1 - \frac{2}{L_2 + \alpha} \alpha\Big)^\tau}{1 - \frac{2}{L_2 + \alpha}  \alpha - \frac{L_2 - \alpha}{L_2 + \alpha}} \Bigg] + L_2 \Delta.
\end{aligned}
\label{eq:15}
\end{equation}
\label{prop:2}
\end{proposition}
\noindent\textbf{Proof.} Using the triangle inequality, we have:
\begin{equation}
\small
\begin{aligned}
\Big\| \frac{\partial \mathcal{L}(\mathbf{R}_t^j, \mathbf{H}_t^\tau)}{\partial \mathbf{R}_t^j} - \nabla F(\mathbf{R}_t^j) \Big\| 
&\leq \Big\| \frac{\partial \mathcal{L}(\mathbf{R}_t^j, \mathbf{H}_t^\tau)}{\partial \mathbf{R}_t^j} - \frac{\partial \mathcal{L}(\mathbf{R}_t^j, \mathbf{H}_{t,j}^\tau)}{\partial \mathbf{R}_t^j} \Big\| \\
&\quad + \Big\| \frac{\partial \mathcal{L}(\mathbf{R}_t^j, \mathbf{H}_{t,j}^\tau)}{\partial \mathbf{R}_t^j} - \nabla F(\mathbf{R}_t^j) \Big\|.
\end{aligned}
\label{eq:16}
\end{equation}
With the help of Assumptions~\ref{ass 2} and~\ref{ass 4}, the following result holds:
\begin{scriptsize}
\begin{equation} 
\begin{aligned}
\Big\| \frac{\partial \mathcal{L}(\mathbf{R}_t^j, \mathbf{H}_t^\tau)}{\partial \mathbf{R}_t^j} - \nabla F(\mathbf{R}_t^j) \Big\| 
&\leq L_2 \Delta + \Big\| \frac{\partial \mathcal{L}(\mathbf{R}_t^j, \mathbf{H}_{t,j}^{\tau})}{\partial \mathbf{R}_t^j} - \nabla F(\mathbf{R}_t^j) \Big\|.
\end{aligned}
\label{eq:17}
\end{equation}
\end{scriptsize}
Plugging in~\eqref{eq:2} into \eqref{eq:17} yields \eqref{eq:15}. Thus, we complete the proof of Proposition~\ref{prop:2}.

Then, we restate some useful lemmas of \cite{ghadimi2016mini} to deal with the proximal operator.
\begin{lemma}[Lemma 1 in \cite{ghadimi2016mini}]
Let $\mathbf{R}_t^{\tau} = \arg\min_\mathbf{R} \bigg\{ \left\langle \frac{\partial L(\mathbf{R}_t^{\tau-1}, \mathbf{H}_t^\tau)}{\partial \mathbf{R}_t^{\tau-1}},\mathbf{R} \right\rangle + h(\mathbf{R}) + \frac{1}{\eta^\prime} \cdot \frac{1}{2} \|\mathbf R - \mathbf{R}_t^{\tau-1}\|^2 \bigg\}$ \ and \ $\tilde{\mathcal{G}}^{\tau-1}_t = \frac{1}{\eta^{\prime}} \left(\mathbf{R}_t^{\tau-1} - \mathbf{R}_t^\tau\right)$. For all $\tau \geq 1$, we have 
\begin{equation}
\small
\left\langle \frac{\partial \mathcal{L}(\mathbf{R}_t^{\tau-1}, \mathbf{H}_t^\tau)}{\partial \mathbf{R}_t^{\tau-1}}, \tilde{\mathcal{G}}_t^{\tau-1} \right\rangle 
\geq \|\tilde{\mathcal{G}}_t^{\tau-1}\|^2 + \frac{1}{\eta^\prime} \big(h(\mathbf{R}_t^\tau) - h(\mathbf{R}_t^{\tau-1}) \big).
\label{eq:18}
\end{equation}
\label{lemma:4}
\end{lemma}
\begin{lemma}[Lemma 2 in \cite{ghadimi2016mini}]
Define $\left(\mathbf{R}_t^{\tau}\right)^+ = \arg\min_\mathbf{R} \bigg\{ \left\langle \nabla F(\mathbf{R}_t^{\tau-1}), \mathbf{R} \right\rangle 
+ h(\mathbf{R})$ $+ \frac{1}{\eta'} \cdot$ $\frac{1}{2} \|\mathbf{R} - \mathbf{R}_t^{\tau-1}\|^2 \bigg\}$, and let $\mathcal{G}_t^{\tau-1} = \frac{1}{\eta'} \left(\mathbf{R}_t^{\tau-1} - \left(\mathbf{R}_t^{\tau}\right)^+\right), \quad\tilde{\mathcal{G}}_t^{\tau-1} = \frac{1}{\eta'} \big(\mathbf{R}_t^{\tau-1}$ $- \mathbf{R}_t^{\tau}\big)$,
we have
\begin{equation}
\|\mathcal{G}_t^{\tau-1} - \tilde{\mathcal{G}}_t^{\tau-1}\| \leq \|\nabla F(\mathbf{R}_t^{\tau-1}) - \frac{\partial L(\mathbf{R}_t^{\tau-1}, \mathbf{H}_t^\tau)}{\partial \mathbf{R}_t^{\tau-1}}\|.
\label{eq:19}
\end{equation}
\label{lemma:5}
\end{lemma}
\begin{theorem}[Restatement of Theorem \ref{thm1}]
Under Assumptions~\ref{ass 1}--\ref{ass 5}, define $\alpha := -L + \mathcal{V}$, and $\mathcal{G}_t^j = \frac{1}{\eta'} \left(\mathbf{R}_t^j - \mathbf{R}_t^{j+1}\right), \quad 0 \leq j \leq \tau-2$, choose step size $\eta$ to be $\frac{2}{L_2 + \alpha}$, $\eta'$ to be $\frac{1}{6L_0}$, and suppose $\alpha < L_2$, $h(\mathbf{R}^{\tau-1}_t) \leq h(\mathbf{R}^0_t),\forall t$, we have:
\begin{equation}
\frac{1}{\tau T} \sum_{t=0}^{T-1} \sum_{j=1}^{\tau-1} \|\mathcal{G}_t^j\|^2 \leq \mathcal{O}\left(\frac{1}{T}\right).
\label{eq:20}
\end{equation}
\end{theorem}

\noindent\textbf{Proof.} According to the above Lemma~\ref{lemma:2}, the function \(\nabla F(R)\) is \(L_0\)-Lipschitz. Let $\tilde{\mathcal{G}}_t^\tau = \frac{1}{\eta^\prime} \left(\mathbf{R}_t^{\tau-1} - \mathbf{R}_t^\tau \right)$, we have:
\begin{equation}
\scriptsize
\begin{aligned}
&F(\mathbf{R}_t^\tau) 
\leq F(\mathbf{R}_t^{\tau-1}) 
+ \langle \nabla F(\mathbf{R}_t^{\tau-1}), \mathbf{R}_t^\tau - \mathbf{R}_t^{\tau-1} \rangle \\
&+ \frac{L_0}{2} \|\mathbf{R}_t^\tau - \mathbf{R}_t^{\tau-1}\|^2 = F(\mathbf{R}_t^{\tau-1}) - \eta' \left\langle \frac{\partial \mathcal{L}(\mathbf{R}_t^{\tau-1}, \mathbf{H}_t^\tau)}{\partial \mathbf{R}_t^{\tau-1}}, \tilde{\mathcal{G}}_t^{\tau-1} \right\rangle \\
& + \eta' \left\langle \frac{\partial \mathcal{L}(\mathbf{R}_t^{\tau-1}, \mathbf{H}_t^\tau)}{\partial \mathbf{R}_t^{\tau-1}} - \nabla F(\mathbf{R}_t^{\tau-1}), \tilde{\mathcal{G}}_t^{\tau-1} \right\rangle
+ \frac{\eta'^2 L_0}{2} \|\tilde{\mathcal{G}}_t^{\tau-1}\|^2 \\
\end{aligned}
\end{equation}
\begin{equation}
\small
\begin{aligned}
&\leq F(\mathbf{R}_t^{\tau-1}) - \eta' \|\tilde{\mathcal{G}}_t^{\tau-1}\|^2 - h(\mathbf{R}_t^\tau) + h(\mathbf{R}_t^{\tau-1}) \\
& + \eta' \left\langle \frac{\partial \mathcal{L}(\mathbf{R}_t^{\tau-1}, \mathbf{H}_t^\tau)}{\partial \mathbf{R}_t^{\tau-1}} - \nabla F(\mathbf{R}_t^{\tau-1}), \tilde{\mathcal{G}}_t^{\tau-1} \right\rangle \\
&+ \frac{\eta'^2 L_0}{2} \|\tilde{\mathcal{G}}_t^{\tau-1}\|^2 \, (\mathrm{i}) \leq F(\mathbf{R}_t^{\tau-1}) 
+ \left(\frac{\eta'^2 L_0}{2} - \frac{3\eta'}{4}\right) \|\tilde{\mathcal{G}}_t^{\tau-1}\|^2 \\
&- h(\mathbf{R}_t^\tau) + h(\mathbf{R}_t^{\tau-1}) + \eta' \left\|\frac{\partial \mathcal{L}(\mathbf{R}_t^{\tau-1}, \mathbf{H}_t^\tau)}{\partial \mathbf{R}_t^{\tau-1}} - \nabla F(\mathbf{R}_t^{\tau-1})\right\| \, (\mathrm{ii}),
\end{aligned}
\label{eq:21}
\end{equation}
\normalsize
\noindent where $(i)$ holds by the above Lemma~\ref{lemma:4}, and $(ii)$ holds by the Cauchy inequality and the basic inequality.

Let $\mathcal{F}(\mathbf{R}) = F(\mathbf{R}) + h(\mathbf{R})$, plugging the result in Proposition~\ref{prop:2} into~\eqref{eq:19}, we have:
\begin{equation}
\small
\begin{aligned}
\mathcal{F}(\mathbf{R}_t^\tau) &\leq F(\mathbf{R}_t^{\tau-1}) + h(\mathbf{R}_t^{\tau-1}) 
+ \left(\frac{\eta'^2 L_0}{2} - \frac{3\eta'}{4}\right) \|\tilde{\mathcal{G}}_t^{\tau-1}\|^2 \\
&\quad + \eta' \left\| \frac{\partial \mathcal{L}(\mathbf{R}_t^{\tau-1}, \mathbf{H}_t^\tau)}{\partial \mathbf{R}_t^{\tau-1}} - \nabla F(\mathbf{R}_t^{\tau-1}) \right\| \\
&\leq F(\mathbf{R}_t^{\tau-1}) + h(\mathbf{R}_t^{\tau-1}) 
+ \left(\frac{\eta'^2 L_0}{2} - \frac{3\eta'}{4}\right) \|\tilde{\mathcal{G}}_t^{\tau-1}\|^2 \\
&\quad + \eta'(L_2 + \frac{L_2^2}{\alpha}) \left[\left(\frac{L_2 - \alpha}{L_2 + \alpha}\right)^\tau \Delta \right] \\
&\quad  + \eta' L_1 \Big[\frac{L_2 \left(1 - \frac{2}{L_2 + \alpha} \right)^\tau}{\alpha} 
+ \frac{2}{L_2 + \alpha} \left(\frac{L_2 L_4}{\alpha} + L_3\right) \\
&\quad \Delta  \frac{\left(1 - \frac{2}{L_2 + \alpha} \cdot \alpha\right)^\tau}{1 - \frac{2}{L_2 + \alpha} \cdot \alpha - \frac{L_2 - \alpha}{L_2 + \alpha}} \Big] + \eta' L_2 \Delta.
\end{aligned}
\label{eq:22}
\end{equation}
According to Lemma~\ref{lemma:5}, the difference between $\tilde{\mathcal{G}}_t^{\tau-1}$ and $\mathcal{G}_t^{\tau-1}$ are bounded, we have:
\begin{equation}
\begin{aligned}
&\|\mathcal{G}_t^{\tau-1}\|^2 
\leq 2\|\tilde{\mathcal{G}}_t^{\tau-1}\|^2 + 2\|\tilde{\mathcal{G}}_t^{\tau-1} -\mathcal{G}_t^{\tau-1}\|^2 \\
&\leq 2\|\tilde{\mathcal{G}}_t^{\tau-1}\|^2 + 2\left\| \frac{\partial L(\mathbf{R}_t^{\tau-1}, \mathbf{H}_t^\tau)}{\partial \mathbf{R}_t^{\tau-1}} - \nabla F(\mathbf{R}_t^{\tau-1}) \right\|^2 \\
&\leq 2\|\tilde{\mathcal{G}}_t^{\tau-1}\|^2 
+ 4 (L_2 + \frac{L_2^2}{\alpha})^2 
\left[\left(\frac{L_2 - \alpha}{L_2 + \alpha}\right)^\tau \Delta \right]^2 \\
&\quad + 4L_1^2 
\Bigg[
L_2 \left(1 - \frac{2}{L_2 + \alpha} \right)^\tau
\frac{1}{\alpha} 
+ \frac{2}{L_2 + \alpha} 
\Big(\frac{L_2 L_4}{\alpha} + L_3\Big) \\
&\quad \Delta \frac{\left(1 - \frac{2}{L_2 + \alpha} \cdot \alpha\right)^\tau}
{1 - \frac{2}{L_2 + \alpha} \cdot \alpha - \frac{L_2 - \alpha}{L_2 + \alpha}}
\Bigg]^2 + 4L_2^2 \Delta^2.
\end{aligned}
\label{eq:23}
\end{equation}
Thus, we have:
\begin{equation}
\small
\begin{aligned}
&-\|\mathcal{G}_t^{\tau-1}\|^2 
\leq -\frac{1}{2}\|\tilde{\mathcal{G}}_t^{\tau-1}\|^2 
+ 2 (L_2 + \frac{L_2^2}{\alpha})^2 \Bigg[\left(\frac{L_2 - \alpha}{L_2 + \alpha}\right)^\tau \Delta \Bigg]^2\\
&+ 2L_1^2 \Bigg[
L_2 \left(1 - \frac{2}{L_2 + \alpha}\right)^\tau
\frac{1}{\alpha} + \frac{2}{L_2 + \alpha} \\
&\left(\frac{L_2 L_4}{\alpha} + L_3\right) \Delta
\frac{\left(1 - \frac{2}{L_2 + \alpha} \cdot \alpha\right)^\tau}
{1 - \frac{2}{L_2 + \alpha} \cdot \alpha - \frac{L_2 - \alpha}{L_2 + \alpha}}
\Bigg]^2 + 2L_2^2 \Delta^2.
\end{aligned}
\label{eq:24}
\end{equation}
Plugging (\ref{eq:24}) into (\ref{eq:22}), we have:
\begin{equation}
\small
\begin{aligned}
\mathcal{F}(\mathbf{R}_t^\tau) &\leq F(\mathbf{R}_t^{\tau-1}) + h(\mathbf{R}_t^{\tau-1}) 
+ \left(\frac{\eta'^2 L_0}{4} - \frac{3\eta'}{8}\right) \|{\mathcal{G}}_t^{\tau-1}\|^2 \\
&\quad + 2\left(\frac{7\eta'}{4} - \frac{\eta'^2 L_0}{2}\right)(L_2 + \frac{L_2^2}{\alpha})^2 \left[\left(\frac{L_2 - \alpha}{L_2 + \alpha}\right)^\tau \Delta \right]^2\\
&\quad  + 2\left(\frac{7\eta'}{4} - \frac{\eta'^2 L_0}{2}\right) L_1^2 \Bigg[\frac{L_2 \left(1 - \frac{2}{L_2 + \alpha}\right)^\tau}{\alpha} \nonumber\\
&\quad + \frac{2}{L_2 + \alpha} \left(\frac{L_2 L_4}{\alpha} + L_3\right) \Delta  \frac{\left(1 - \frac{2}{L_2 + \alpha}  \alpha\right)^\tau}{1 - \frac{2}{L_2 + \alpha} \cdot \alpha - \frac{L_2 - \alpha}{L_2 + \alpha}} \Bigg]^2 \\
&\quad + 2\left(\frac{7\eta'}{4} - \frac{\eta'^2 L_0}{2}\right) L_2^2 \Delta^2.
\end{aligned}
\label{eq:25}
\end{equation}
Based on the $L_0$-smoothness of $F(\mathbf{R})$ established in Lemma~\ref{lemma:2}, for $j \geq 1$, we have:
\begin{equation}
\small
\begin{aligned}
F(\mathbf{R}_t^{j+1}) &\leq F(\mathbf{R}_t^j) - (\frac{\eta'}{2}-{\eta'^2 L_0}) \|\nabla F(\mathbf{R}_t^j)\|^2 \\
&\quad +  \Big\|\frac{\partial L(\mathbf{R}_t^{j}, \mathbf{H}_t^\tau)}{\partial \mathbf{R}_t^{j}}-\nabla F(\mathbf{R}_t^j)\Big\|^2\cdot (\frac{\eta'}{2}+{\eta'^2 L_0})  \\
&\leq F(\mathbf{R}_t^j) - (\frac{\eta'}{2}-{\eta'^2 L_0}) \|\nabla F(\mathbf{R}_t^j)\|^2\\
&\quad + \left(\eta' + 2\eta'^2 L_0\right)(L_2 + \frac{L_2^2}{\alpha})^2 \left(\frac{L_2 - \alpha}{L_2 + \alpha}\right)^{2\tau} \Delta ^2 \\
&\quad + \left(\eta' + 2\eta'^2 L_0\right)L_1^2 \Bigg[\frac{L_2 \left(1 - \frac{2}{L_2 + \alpha}\right)^\tau}{\alpha} + \frac{2}{L_2 + \alpha} \\
&\quad\left(\frac{L_2 L_4}{\alpha} + L_3\right) \Delta \frac{\left(1 - \frac{2}{L_2 + \alpha} \cdot \alpha\right)^\tau}{1 - \frac{2}{L_2 + \alpha} \cdot \alpha - \frac{L_2 - \alpha}{L_2 + \alpha}}\Bigg]^2 \\
&\quad + \left(\eta' + 2\eta'^2 L_0\right)L_2^2 \Delta^2.
\end{aligned}
\label{eq:26}
\end{equation}
Then, with $\eta' = \frac{1}{6L_0}$, telescoping \eqref{eq:26} over $j$ from $0$ to $\tau - 2$ yields:
\begin{equation}
\small
\begin{aligned}
F(\mathbf{R}_t^{\tau-1}) &\leq F(\mathbf{R}_t^0) 
- \frac{1}{18L_0} \sum_{j=0}^{\tau-2} \|\nabla F(\mathbf{R}_t^j)\|^2 
+ \frac{2}{9L_0} (\tau-1) \\
&\quad \Bigg\{(L_2 + \frac{L_2^2}{\alpha})^2 \left(\frac{L_2 - \alpha}{L_2 + \alpha}\right)^{2\tau} \Delta ^2 + L_1^2 \\
&\quad\bigg[\frac{L_2 \left(1 - \frac{2}{L_2 + \alpha}\right)^\tau}{\alpha} 
+ \frac{2}{L_2 + \alpha} \left(\frac{L_2 L_4}{\alpha} + L_3\right) \\
&\quad \Delta  \frac{\left(1 - \frac{2}{L_2 + \alpha}  \alpha\right)^\tau}{1 - \frac{2}{L_2 + \alpha} \cdot \alpha - \frac{L_2 - \alpha}{L_2 + \alpha}} \bigg]^2 \Bigg\}   + \frac{2}{9L_0} (\tau-2) L_2^2\Delta^2.
\end{aligned}
\label{eq:27}
\end{equation}
Plugging \eqref{eq:27} into \eqref{eq:25}, and $h(\mathbf{R}^{\tau-1}_t) \leq h(\mathbf{R}^0_t)$ yields:
\begin{equation}
\small
\begin{aligned}
\mathcal{F}(\mathbf{R}_t^\tau) &\leq F(\mathbf{R}_t^0) + h(\mathbf{R}_t^0) 
- \frac{1}{18L_0} \|\mathcal{G}_t^{\tau-1}\|^2 -\frac{1}{18L_0} \sum_{j=0}^{\tau-2} \|\mathcal{G}_t^j\|^2 \\
&+ \left[\frac{5}{9L_0} + \frac{2}{9L_0} (\tau-1)\right]\Bigg\{ (L_2 + \frac{L_2^2}{\alpha})^2 \left(\frac{L_2 - \alpha}{L_2 + \alpha}\right)^{2\tau} \Delta^2 
\end{aligned}
\end{equation}
\begin{equation}
\small
\begin{aligned}
&+ L_1^2 \bigg[\frac{L_2 \left(1 - \frac{2}{L_2 + \alpha}\right)^\tau}{\alpha} + \frac{2}{L_2 + \alpha} \left(\frac{L_2 L_4}{\alpha} + L_3\right)\Delta\\
&\quad \frac{\left(1 - \frac{2}{L_2 + \alpha}  \alpha\right)^\tau}{1 - \frac{2}{L_2 + \alpha} \alpha - \frac{L_2 - \alpha}{L_2 + \alpha}} \bigg]^2 \Bigg\}+ \left(\frac{5}{9L_0} + \frac{2}{9L_0} (\tau-2)\right) L_2^2 \Delta^2.
\end{aligned}
\label{eq:28}
\end{equation}
Telescoping \eqref{eq:28} over $t$ from $0$ to $T-1$ yields:
\begin{equation}
\small
\begin{aligned}
&\frac{1}{\tau T} \sum_{t=0}^{T-1} \sum_{j=0}^{\tau-1} \|\mathcal{G}_t^j\|^2 \leq 
18L_0\frac{\mathcal{F}(\mathbf{R}_0^0) - \inf\limits_{\mathbf{R}} \mathcal{F}(\mathbf{R})}{\tau \cdot T} \\
&\quad + \frac{1}{\tau } \left[\frac{5}{9L_0} + \frac{2}{9L_0} (\tau - 1)\right] \Bigg\{\quad (L_2 + \frac{L_2^2}{\alpha})^2 \left(\frac{L_2 - \alpha}{L_2 + \alpha}\right)^{2\tau} \Delta^2 \\
&\quad + L_1^2 \bigg[ \frac{L_2 (1 - \frac{2}{L_2 + \alpha})^\tau}{\alpha} \quad + \frac{2}{L_2 + \alpha} \left(\frac{L_2 L_4}{\alpha} + L_3\right)\Delta \cdot
\end{aligned}
\end{equation}
\begin{equation}
\small
\begin{aligned}
&\quad \frac{(1 - \frac{2}{L_2 + \alpha}\quad \cdot \alpha)^\tau}
{1 - \frac{2}{L_2 + \alpha} \cdot \alpha - \frac{L_2 - \alpha}{L_2 + \alpha}} 
\bigg]^2 \Bigg\} + \frac{1}{\tau } \left[
\frac{5}{9L_0} + \frac{2}{9L_0} (\tau - 2)
\right] L_2^2 \Delta^2.
\end{aligned}
\label{eq:29}
\end{equation}
Thus, we complete the proof of Theorem 1.

\end{document}